%% file: main.tex
\documentclass{article}

\usepackage{PRIMEarxiv}

\usepackage[utf8]{inputenc} 
\usepackage[T1]{fontenc}    
\usepackage{hyperref}       
\usepackage{url}            
\usepackage{booktabs}       
\usepackage{amsfonts}       
\usepackage{nicefrac}       
\usepackage{microtype}      
\usepackage{lipsum}
\usepackage{threeparttable}
\usepackage{fancyhdr}       
\usepackage{graphicx}       
\graphicspath{{media/}}     

\usepackage{graphicx}%
\usepackage{makecell}
\usepackage{multirow}%
\usepackage{amsthm}%
\usepackage{mathrsfs}%
\usepackage[title]{appendix}%
\usepackage{xcolor}%
\usepackage{textcomp}%
\usepackage{tabularx}
\usepackage[most]{tcolorbox}
\usepackage{booktabs}
\usepackage{multirow}
\usepackage[table]{xcolor}
\usepackage{multirow}
\usepackage{pifont}
\usepackage{tikz}
\usepackage{pgfplots}
\pgfplotsset{compat=1.18}
\usepackage{subcaption} 
\usepackage{pgfplotstable}
\usepackage{wrapfig}
\usepackage{siunitx}
\usepackage{threeparttable}
\usepackage{manyfoot}%
\usepackage{booktabs}%
\usepackage{algorithm}%
\usepackage{algorithmicx}%
\usepackage{algpseudocode}%
\usepackage{longtable}
 \usepackage{array}
\usepackage{tcolorbox}
\usepackage{listings}%
\usepackage{array}

\newcommand{\cmark}{\ding{51}}
\newcommand{\xmark}{\ding{55}}
\newcommand{\pmrk}{\(\circ\)} 
\newcommand{\na}{---}
\newcolumntype{P}[1]{>{\centering\arraybackslash}p{#1}}
\newcolumntype{M}[1]{>{\centering\arraybackslash}m{#1}}
\newcolumntype{L}[1]{>{\raggedright\arraybackslash}m{#1}} 
\newcolumntype{C}[1]{>{\centering\arraybackslash}m{#1}}   
\newcolumntype{R}[1]{>{\raggedleft\arraybackslash}m{#1}} 

\newcolumntype{x}[1]{>{\raggedright\arraybackslash}p{#1}} 
\AtBeginDocument{}

\definecolor{SmallVLMBlue}{rgb}{0.90,0.95,1.0}
\definecolor{LargeVLMYellow}{rgb}{1.0,0.97,0.85}

\definecolor{SmallVLMBlue}{rgb}{0.90,0.95,1.0}
\definecolor{LargeVLMPeach}{rgb}{1.0,0.92,0.88}

\definecolor{SmallVLMTeal}{rgb}{0.85,0.95,0.95}
\definecolor{LargeVLMBeige}{rgb}{0.96,0.94,0.88}

\definecolor{SmallVLMCyan}{rgb}{0.88,0.97,1.0}
\definecolor{LargeVLMSand}{rgb}{0.98,0.95,0.85}

\definecolor{SmallVLMSoftBlue}{rgb}{0.90,0.95,1.0}
\definecolor{LargeVLMCoral}{rgb}{1.0,0.88,0.88}

\title{ViLBias: Detecting and Reasoning about Bias in Multimodal Content}

\author{
  Shaina Raza\thanks{shaina.raza@torontomu.ca} \\
  Vector Institute \\
  Toronto, ON, Canada \\
  \And
  Caesar Saleh \\
  Vector Institute \\
  Toronto, ON, Canada \\
  \\
  \And
  Azib Farooq \\
  University of Cincinnati \\
  Cincinnati, OH, USA \\
  \And
  Emrul Hasan \\
  Vector Institute \\
  Toronto, ON, Canada \\
  \\
  \And
  Franklin Ogidi \\
  Vector Institute \\
  Toronto, ON, Canada \\
  \\
  \And
  Haad Zahid \\
  LUMS \\
  Punjab, Pakistan \\
  \\
  \And
  Maximus Powers \\
  Arizona State University \\
  Arizona, United States \\
  \\
  \And
  Marcelo Lotif \\
  Vector Institute \\
  Toronto, ON, Canada \\
  \\
  \And
  Anam Zahid \\
 ITU \\
  Punjab, Pakistan \\
  \\
  \And
 Karanpal Sekhon \\
  Vector Institute \\
  Toronto, ON, Canada \\
  \\
  \And
 Veronica Chatrath \\
  Vector Institute \\
  Toronto, ON, Canada \\
  \\
  \And
   Roya Javedi \\
  Vector Institute \\
  Toronto, ON, Canada \\
  \\
  \And
  Vahid Reza Khazaie \\
  Vector Institute \\
  Toronto, ON, Canada \\
  \\
    \And
  Zhenyu Yu \\
  Universiti Malaya \\
  Kuala Lumpur, Malaysia \\
}

\begin{document}
\maketitle

\input{section/0-abstract}


\begin{figure}
  \includegraphics[width=1\linewidth]{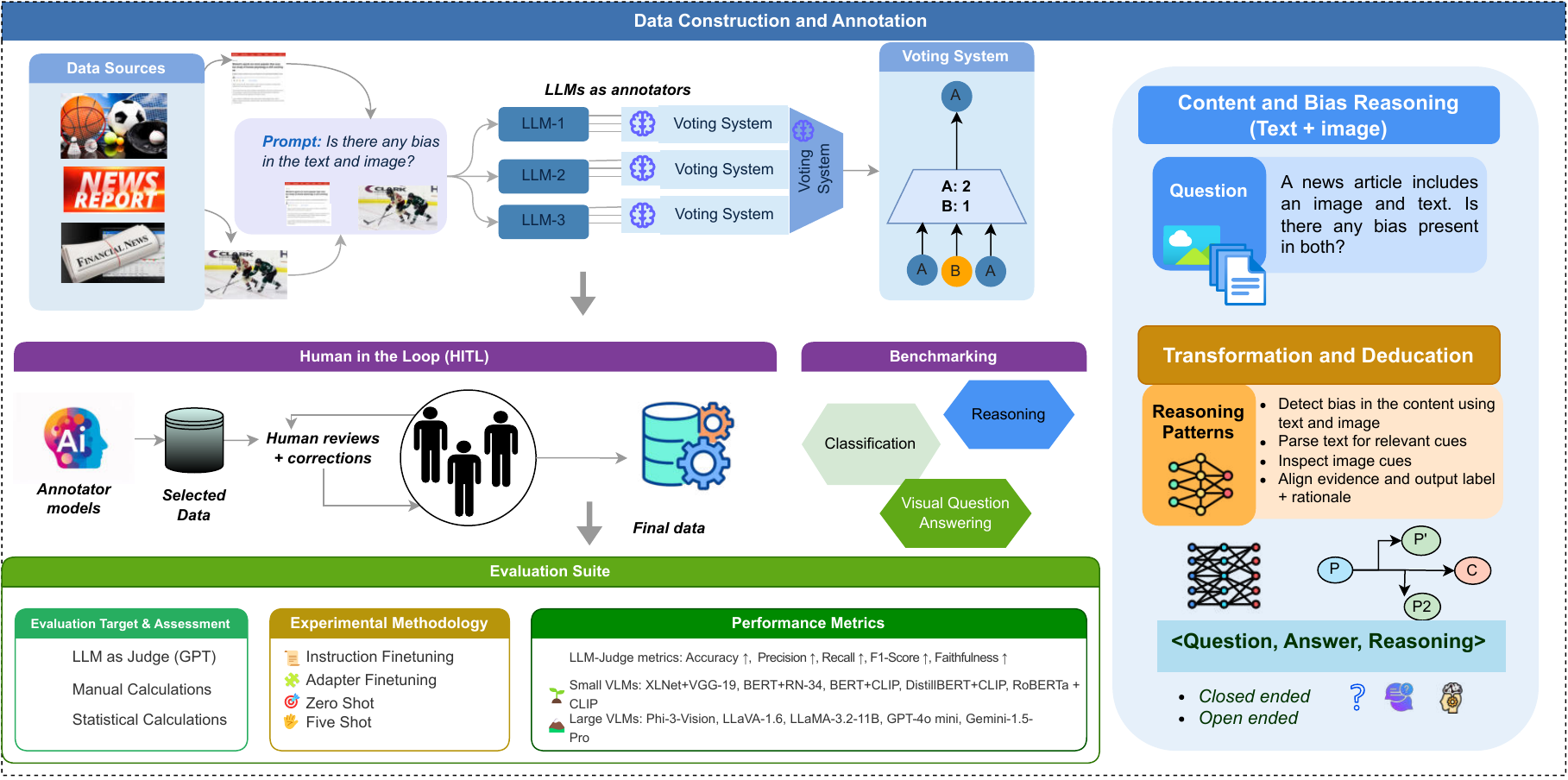}
  \caption{\textbf{\textsc{V}iLBias} Framework. The pipeline comprises three main stages: (1) Data Collection and Preprocessing, (2) Automated and Human-Annotated Labeling, and (3) Final Evaluation.}
  \label{fig:teaser}
\end{figure}

\maketitle

\renewcommand\thefootnote{}

\renewcommand\thefootnote{\fnsymbol{footnote}}
\setcounter{footnote}{1}
\input{section/1-introduction}

\input{section/2-Related}
\input{section/3-Methodology}
\input{section/4-Results} 
\input{section/6-Discussion}
\input{section/7-Conclusion}

\bibliographystyle{unsrt}  
\bibliography{main}


\input{section/8-Appendix}
\input{section/haad_eda}

\end{document}

%% file: section/0-abstract.tex
\begin{abstract}
Detecting bias in multimodal news requires models that reason over text–image pairs, not just classify text. In response, we present ViLBias, a VQA-style benchmark and framework for detecting and reasoning about bias in multimodal news. The dataset comprises 40,945 text–image pairs from diverse outlets, each annotated with a bias label and concise rationale using a two-stage LLM-as-annotator pipeline with hierarchical majority voting and human-in-the-loop validation. We evaluate Small Language Models (SLMs), Large Language Models (LLMs), and Vision–Language Models (VLMs) across closed-ended classification and open-ended reasoning (oVQA), and compare parameter-efficient tuning strategies. Results show that incorporating images alongside text improves detection accuracy by 3–5\%, and that LLMs/VLMs better capture subtle framing and text–image inconsistencies than SLMs. Parameter-efficient methods (LoRA/QLoRA/Adapters) recover 97–99\% of full fine-tuning performance with <5\% trainable parameters. For oVQA, reasoning accuracy spans 52–79\% and faithfulness 68–89\%, both improved by instruction tuning; closed accuracy correlates strongly with reasoning (r = 0.91). ViLBias offers a scalable benchmark and strong baselines for multimodal bias detection and rationale quality.
The data and code is available here \footnote{Anonymous due to double blind review}.

\end{abstract}

%% file: section/1-introduction.tex
\section{Introduction}
Recent advances in natural language processing (NLP) and multimodal learning, exemplified by the OpenAI GPT series \cite{openai2024gpt4o} and the Meta LLaMA series \cite{meta2024llama}, have substantially improved the analysis and interpretation of textual and visual content. However, despite their capacity to process large volumes of information, these systems can also inherit and amplify biases present in both modalities. In multimodal settings, bias extends beyond language to include visual strategies such as image selection, cropping, staging, and emotion-provoking imagery .

In this work, we define \textit{bias} as directional framing in text, image, or their interaction that skews salience or interpretation (e.g., loaded wording, selective emphasis, image selection/cropping/staging, cross-modal mismatch). Our focus is news media, and we distinguish framing bias from misinformation. The topic of  multimodal fake news detection has received substantial attention \cite{alam-etal-2022-survey}, systematic bias detection remains comparatively underexplored, partly due to its inherent subjectivit, which underscores the need for targeted research in this domain.


Existing benchmarks for news and social media bias remain limited, with most focusing on text-only signals. Resources such as MBIB \cite{wessel2023introducing}, BABE \cite{spinde2022neural}, and related corpora \cite{rodrigo2024systematic} emphasize textual bias. Multimodal datasets (e.g., NewsBag \cite{NewsBag2020dataset}, MMFakeBench \cite{liu2024mmfakebench}, FakeNewsNet \cite{shu_fakenewsnet_2020}) primarily address misinformation rather than \emph{framing} bias. Current research shows three gaps: (i) most benchmarks target fake news/misinformation rather than framing bias; when bias is considered, resources are predominantly \emph{text-only} and multimodal bias benchmarks are scarce; (ii) most resources are \emph{label-only}, lacking rationales or evidence, which precludes evaluation of \emph{reasoning quality}; and (iii) scalable, audited annotation protocols (e.g., LLM-as-annotator with human validation) are uncommon, limiting dataset size and consistency. These gaps motivate benchmarks that integrate textual and visual dimensions of bias and evaluate both \emph{closed-ended} classification and \emph{open-ended} reasoning (visual question answering (VQA)-style). A comparison with state-of-the-art (SOTA) bias-detection benchmarks is provided in Table~\ref{tab:bias_sota_benchmarks}.

\begin{table}[t]
\scriptsize
\centering
\caption{Bias and related multimodal (MM) benchmarks. 
Legend: \cmark=yes, \xmark=no, \pmrk=partial; 
Mod.=modality (T=Text, I=Image); 
Annot.=annotation source (AI/LLM, H=Human); 
Rat.=rationale; MM=multimodal checks; Reason.=reasoning.}
\renewcommand{\arraystretch}{1.0}
\setlength{\tabcolsep}{2pt}
\resizebox{\linewidth}{!}{%
\begin{tabular}{@{}L{0.20\columnwidth} C{0.12\columnwidth} C{0.11\columnwidth} 
C{0.11\columnwidth} C{0.08\columnwidth} C{0.10\columnwidth} C{0.10\columnwidth}@{}}
\toprule
\textbf{Benchmark} & \textbf{Mod.} & \textbf{Focus} & \textbf{Annot.} & \textbf{Rat.} & \textbf{MM chk.} & \textbf{Reason.} \\
\midrule
BiasLab \cite{solaiman2025biaslab} & T & Bias & H & \cmark & \na & \pmrk \\
BIASsist \cite{noh2025biassist}    & T & Bias & AI/LLM & \cmark & \na & \cmark \\
MDAM3 \cite{xu2025mdam3}           & T+I & Misinfo & AI+H & \xmark & \pmrk & \xmark \\
MMFakeBench \cite{liu2024mmfakebench} & T+I & Misinfo & AI+H & \xmark & \cmark & \pmrk \\
IndiTag \cite{lin2024inditag}      & T & Bias  & AI/LLM & \pmrk & \na & \pmrk \\
VERITE \cite{papadopoulos2024verite} & T+I & Misinfo & H & \xmark & \cmark & \xmark \\
MBIB \cite{wessel2023introducing}  & T & Bias & H & \pmrk & \na & \xmark \\
BABE \cite{spinde2022neural}       & T & Bias & H & \cmark & \na & \xmark \\
BASIL \cite{fan2019plain}          & T & Bias & H & \cmark & \na & \xmark \\
NewsBag \cite{NewsBag2020dataset}  & T+I & Misinfo & H & \xmark & \pmrk & \xmark \\
FakeNewsNet \cite{shu_fakenewsnet_2020} & T+I & Misinfo & AI+H & \xmark & \pmrk & \xmark \\
\textbf{ViLBias (ours)}            & \textbf{T+I} & \textbf{Bias} & \textbf{AI+H} & \cmark & \cmark & \cmark \\
\bottomrule
\end{tabular}
}
\label{tab:bias_sota_benchmarks}
\end{table}

We present \textsc{ViLBias}, a VQA-style multimodal benchmark for news media bias. Each text–image pair is combined with a question, a label (bias presence/type), and a concise reasoning (rationale), enabling evaluation of both prediction and reasoning. \textsc{ViLBias} consists of 40k unique multimodal pairs curated from diverse real-world news sources. Given the scale of the data, annotations are produced through a hybrid strategy that combines LLM-based automated labeling with extensive human review to ensure quality and consistency. The dataset covers multiple bias categories, ranging from overt ideological framing to subtle visual manipulations such as cropping, staging, and emotive imagery. We evaluate a range of state-of-the-art (SoTA) small language models (SLMs)\footnote{We define SLMs as encoder-only pre-ChatGPT architectures \cite{wang2024comprehensive}.}, large language models (LLMs) and vision languahe models (VLMs), assessing their ability to answer multimodal bias-related questions.
 
\paragraph{Contributions} Our work can be summarized into four main contributions:
\begin{enumerate}
    \item We present ViLBias, a VQA-style multimodal benchmark for news media bias.
    \item We introduce BiasCorpus, a dataset of 40k text–image pairs from diverse news sources, each annotated with bias labels and rationales to frame bias detection as a multimodal task.
    \item We design a hybrid annotation pipeline that stabilizes outputs through multi-LLM voting and integrates expert human review to ensure high-quality annotations.
    \item We conduct a systematic evaluation of SLMs, LLMs, and VLMs, highlighting their respective strengths and limitations in multimodal media bias detection.
\end{enumerate}

%% file: section/2-Related.tex
\section{Related Work}
\label{sec:related} 
\begin{figure*}[t]
\centering
\includegraphics[width=0.88\textwidth]{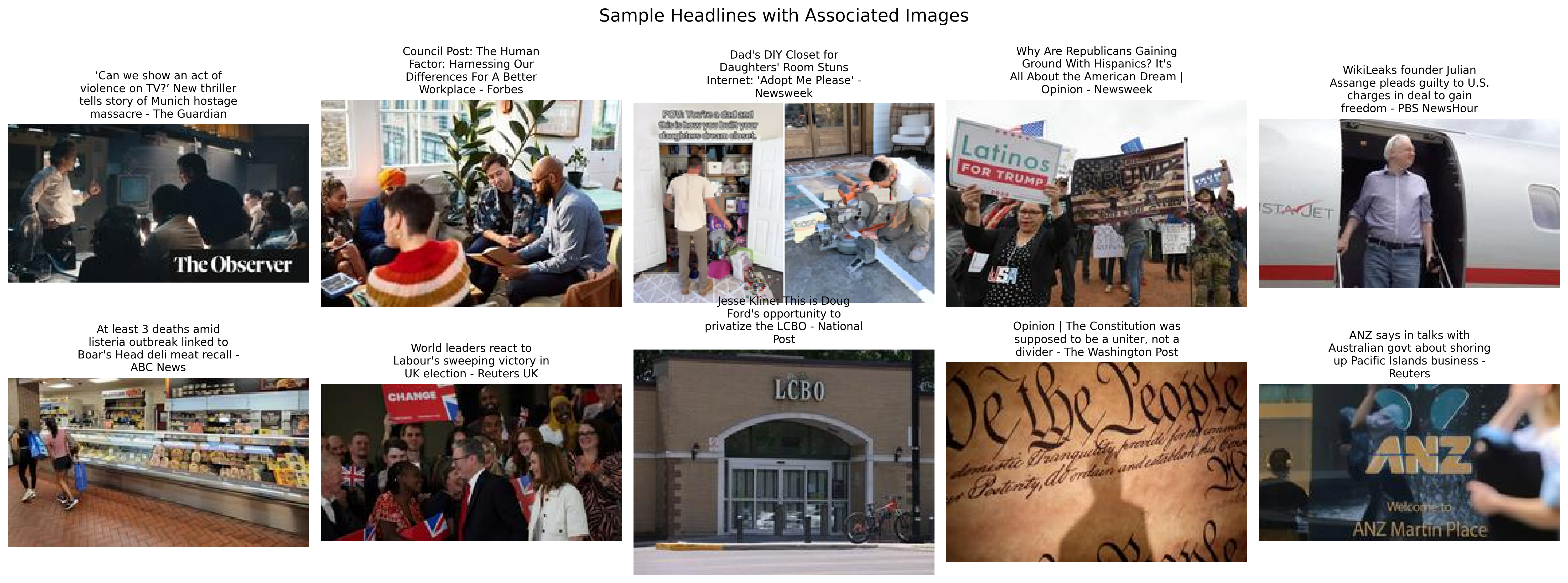}
    \caption{Sample headlines and their associated images. Each image corresponds to a specific news article, illustrating the multimodal structure of the dataset.}
\end{figure*}
The study of fake news can be categorized into disinformation, misinformation  \cite{santos2021misinformation}, and bias news \cite{raza2024unlocking}.  Biased news typically presents information with a particular slant, emphasizing certain viewpoints while downplaying others \cite{raza2024nbias}, whereas fake news or misinformation consists of entirely fabricated content intended to mislead readers. This work primarily focuses on bias detection in news.
Research on bias detection typically uses SLMs for classification and LLMs  for open-ended reasoning.  Prior work indicates that SLMs can be effective on overt lexical biases when fine-tuned for specific tasks \cite{zhang2024tinyllama}, but they struggle with subtler forms such as framing and ideological slant, where LLMs demonstrate stronger generalization and reasoning abilities \cite{prabhumoye2021few,wu2024reasoning}. 

\textbf{Annotation Process}
Multimodal bias detection suffers from limited labeled data. Early efforts relied on crowd workers \cite{tschiatschek2018fake} or experts \cite{wang-2017-liar}, with some attempts at AI-generated labels \cite{gilardi2023chatgptdatacollection}. Manual labeling is costly, slow, and error-prone, prompting interest in AI-assisted approaches \cite{desmond2021increasing}. Recent studies use GPT models for annotation \cite{tan2024large}, leading to hybrid frameworks that combine AI-generated labels with human verification \cite{guan2024language}, while others rely solely on AI annotations \cite{papado2023misinformer}. 

\textbf{Multimodal Bias Detection.}
Bias in news often emerges from text–image interactions, requiring models that capture cross-modal framing \cite{adewumi2024fairness,comito2023multimodal}. Approaches using CLIP-style alignment or contrastive objectives improve detection but are largely aimed at misinformation rather than bias \cite{rahman2022multimodal,cui2024migcl}. Benchmarks reveal pitfalls such as unimodal shortcuts, where text- or image-only models perform well on multimodal tasks \cite{papadopoulos2024verite}. Recent efforts like VLBiasBench expose demographic biases in VLMs, highlighting the need for reasoning-oriented benchmarks \cite{wang2024vlbiasbench}. Empirically, visual cues provide modest gains (approx. 3–5\%) beyond text, underscoring the value of multimodal supervision. A broader review of multimodal fake and bias detection is given in Appendix Table~\ref{tab:multimodal_summary}. 

In this work, we address these gaps by introducing \textbf{ViLBias}, a multimodal benchmark with binary labels and reasonings, annotated via an LLM-as-annotator plus human validation pipeline, and we establish baselines for both closed-ended classification and open-ended (VQA-style) models.

%% file: section/3-Methodology.tex
\section{Methodology}
\label{sec:method}
    \label{fig:headlines}

Our proposed framework, \textbf{\textsc{V}iLBias}, is structured into three core stages, as illustrated in Figure~\ref{fig:teaser}. First, we introduce \textit{BiasCorpus}, which is a systematic collection and preparation of multimodal data relevant to news bias detection. Second, we employ a \textit{Human-In-The-Loop (HITL) BiasAnnotation} process to incorporate two step expert-driven annotations, enhancing the reliability of the dataset. Finally, the curated and annotated data are integrated into \textit{BiasEvalSuite}, a comprehensive evaluation suite designed to benchmark and analyze model performance under diverse bias scenarios. Together, these components provide a robust and transparent pipeline for studying bias in multiple models.

\subsection{Problem Definition}
\label{sec:problem_definition}
The main goal of this work is to assess the ability of language models, specifically VLMs, to detect and reason about media bias. We introduce BiasCorpus, a dataset usable in both classification (close-ended) and VQA-style (open-ended) settings. In the latter, each text–image pair is paired with a query such as ``Does this content exhibit bias? Please explain your reasoning''. Formally, given a textual snippet $\mathcal{T}$, a corresponding visual element $\mathcal{V}$, and a bias-related query $\mathcal{Q}$, the task is defined as
\[
f:\;(\mathcal{T}, \mathcal{V}, \mathcal{Q}) \;\mapsto\; (\mathcal{A}_{\mathrm{cls}}, \mathcal{A}_{\mathrm{reasoning}}),
\]
where $\mathcal{A}_{\mathrm{cls}} \in \{\text{biased}, \text{not biased}\}$ is the categorical prediction and $\mathcal{A}_{\mathrm{reasoning}}$ is an open-ended explanation that justifies the decision by pointing to textual framing (e.g., emotive wording, selective omission) and/or visual cues (e.g., cropping, staging, image sentiment). This joint formulation allows evaluation of both predictive accuracy and reasoning quality.

\subsection{BiasCorpus Curation}
We constructed \textsc{BiasCorpus} by ingesting articles from Google News RSS and official outlet feeds (May~2023–September~2024), de-duplicating by canonical URLs and filtering non-article formats. The sources, spanning CNN, BBC, The New York Times, Al Jazeera, and others as seen in Table~\ref{tab:analysis}, were selected to cover varied geographies, editorial orientations, and audience trust profiles \cite{allsides_media_bias,yougov_trust_media_2024}.
To ensure quality, we removed duplicates, discarded short articles (fewer than twenty sentences), and filtered out multimedia-only items (e.g., videos, slideshows). Each retained article was paired with its first encountered image to establish a consistent multimodal reference. Portions of the text–image pairs overlap with our prior work \footnote{Anonymous due to double blind policy.}, but the present study introduces a distinct labeling protocol based on multi-LLM annotation with human review for robustness. Sample headlines and images are shown in Figure~\ref{fig:headlines}.

\subsection{HITL BiasAnnotation}
\begin{figure}[ht]
    \centering
\includegraphics[width=1\linewidth]{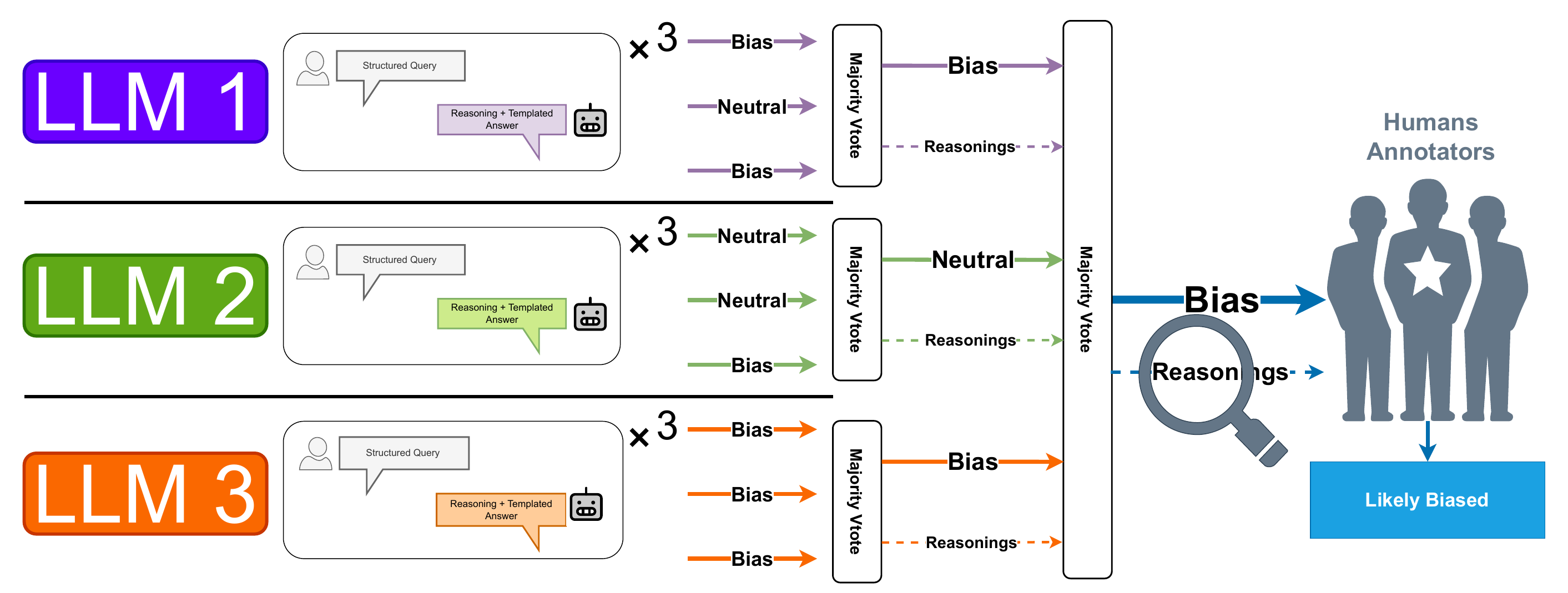}
\caption{LLM-As-Annotators with Human-In-The-Loop (HITL) BiasAnnotation Framework. Each LLM is queried three times, and its responses are majority-voted to remove stochasticity. These majority-voted labels are taken from three LLMs and then majority-voted (across LLMs) again to produce a final label, which is reviewed---along with LLM reasoning---by human annotators.}
\label{fig:annotate}
\end{figure}

To efficiently annotate a large, complex dataset, we integrated automated annotators, using LLM-as-annotators for bulk labeling of text–image pairs. The use of LLMs for annotation has recently been shown to be an effective and scalable strategy \cite{he2023annollm}. The prompt template is provided in Appendix~\ref{app:prompts}. A distinct aspect of our annotation pipeline is a \textit{two-step voting mechanism} designed to maximize reliability while reducing stochastic variability in LLM outputs. The process, as shown in Figure \ref{fig:annotate} operates as follows:\\
(1) \textbf{Step 1 – Stabilizing each LLM:}  
    For each model, the same item is queried three times. Since LLMs are inherently stochastic \cite{bender2021dangers}, the responses may differ. A majority vote is applied over the three responses to obtain one stable annotation per model.\\
(2) \textbf{Step 2 – Consensus across models:}  
    After Step~1, each LLM contributes a single annotation. With three LLMs in total, a second majority vote is then applied across their outputs, yielding the final annotation for that item.

In short, each LLM first “settles on its own answer” by voting across three attempts, and then the models collectively “agree” through another vote. This hierarchical voting scheme ensures robustness both \textit{within} and \textit{across} models, providing reliable annotations.
\paragraph{Human-In-The-Loop}
Human oversight remained essential for validating automated outputs and refining the annotation pipeline. We engaged 12 reviewers (three undergraduates, four master’s students, two Ph.D. holders, and three senior researchers) spanning computer science, media studies, political science, and linguistics. Two subject-matter experts led the process using bias identification protocols (Appendix~\ref{app:ann}), ensuring coverage of both linguistic and visual biases as defined by the U.S. Equal Employment Opportunity Commission \cite{eeoc_discrimination}.  
Reviewers assessed all annotations, classifying them as \textit{Acceptable}, \textit{Needs Improvement}, or \textit{Incorrect}, and provided rationales where needed. Their feedback informed iterative prompt and pipeline adjustments, with disagreements resolved through consensus. After refinement, the inter-annotator agreement (IAA) between automated labels and expert judgments, measured with Cohen’s Kappa, reached 0.72, indicating substantial alignment.
Table~\ref{tab:analysis} summarizes dataset schema, key statistics, and observed patterns.

\subsection{Benchmarking Suite}
We evaluate models on both closed-ended classification and open-ended reasoning tasks across three model families (LLMs, SLMs and VLMs) and two accessibility settings (open-ended and closed-ended).  
First, we consider open-source SLMs as text-only baselines, including encoder-only BERT-style transformers, CLIP-style joint encoders, decoder-only models such as GPT-2, and encoder--decoder models such as BART-base, all fine-tuned on our dataset.  
Second, we examine autoregressive decoder-only LLMs, tested under zero-shot, few-shot, and instruction fine-tuning (IFT) regimes. In addition to open-source models, we also evaluate closed-source APIs such as GPT-4o \cite{gpt4o} and Gemini 1.5 Pro \cite{team2023gemini}.  
Third, we extend evaluation to VLMs, which are assessed not only on classification accuracy but also on their ability to generate faithful, open-ended rationales grounded in both textual and visual evidence.  
For classification tasks, we report standard metrics including precision, recall, F1 score, and accuracy. For open-ended reasoning, we employ an LLM-as-a-judge \cite{zheng2023judging} framework to evaluate the quality, faithfulness, and consistency of generated rationales. Prompts for annotation are in Appendix \ref{app:prompts} and for LLM-judge evaluation in Figure \ref{fig:judge-prompt}.

\begin{table}[h]
\footnotesize
\centering
\caption{Dataset analysis summary.}
\renewcommand{\arraystretch}{1.05}
\begin{tabular}{@{}p{0.35\linewidth}p{0.55\linewidth}@{}}
\toprule
\textbf{Metric} & \textbf{Value} \\
\midrule
Rows & 40,945 \\
\midrule
Field names & \texttt{unique\_id}, \texttt{headline}, \texttt{url}, 
\texttt{article\_text}, \texttt{image}, \texttt{multimodal\_label}, \texttt{reasoning} \\
\midrule
Top outlets & \textit{Financial Times} (4,895), \textit{USA TODAY} (3,320), 
\textit{CNN} (3,067), \textit{The Guardian} (2,833), \textit{Forbes} (2,482) \\
\midrule
Multimodal label classes & 2 (\textit{Biased}=22,974; \textit{Not Biased}=17,971) \\
\midrule
Unique images & 40,945 \\
\bottomrule
\end{tabular}
\label{tab:analysis}
\end{table}

\begin{figure}[h]
    \centering
        \centering
        \includegraphics[width=0.9\linewidth]{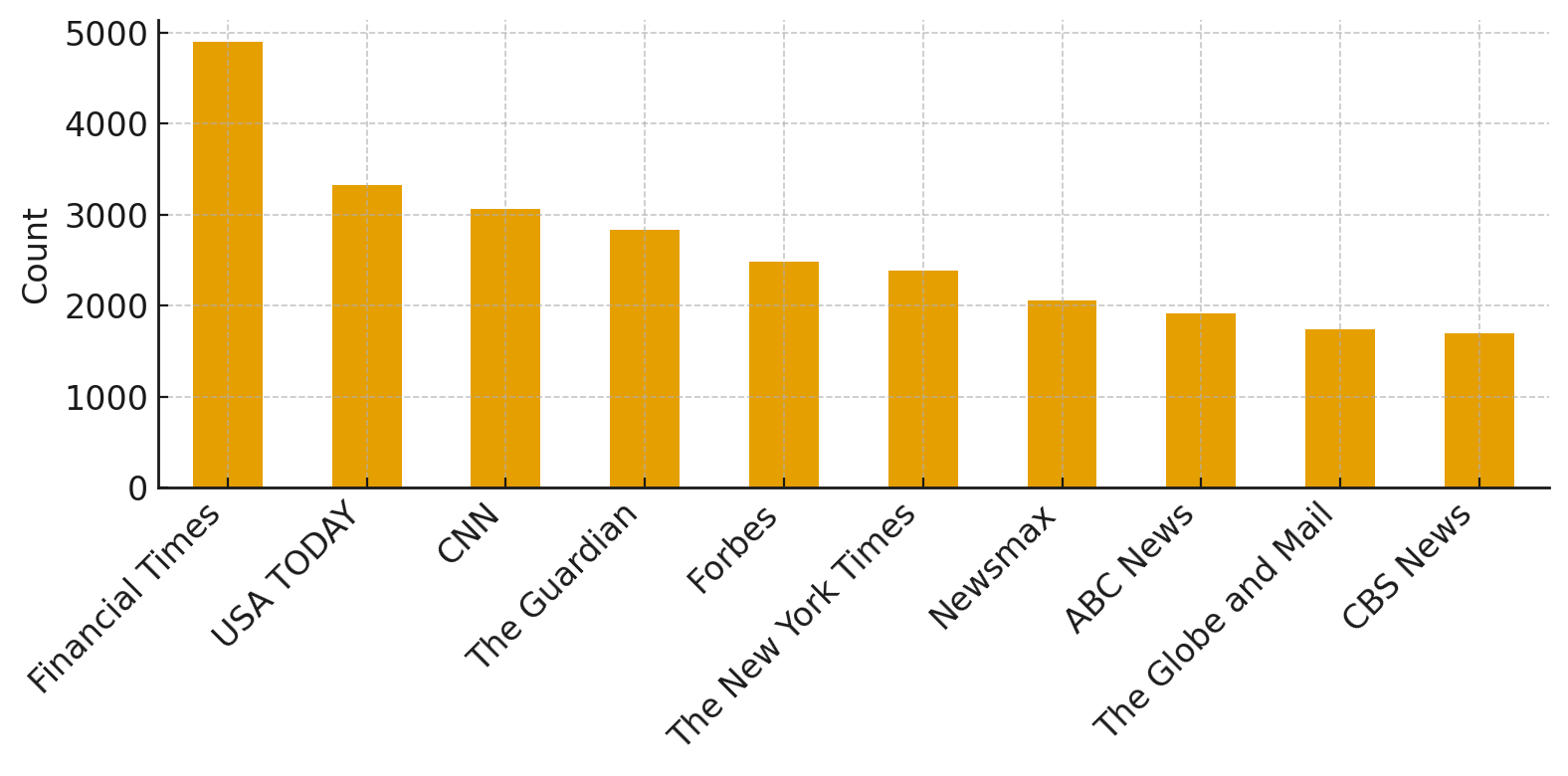}
        \caption{Top 10 outlets}
    \label{fig:eda_minipage}
\end{figure}

%% file: section/4-Results.tex
\section{Results and Discussion}

\subsection{Experimental Setting}
\label{sec:experiments}

\textbf{Compute \& software.}
We run our experiments on NVIDIA A40/A100 Tensor Core GPUs. The stack used Python~3.10, PyTorch~2.0, Hugging Face Transformers, and OpenCV.

\textbf{Baselines and model families.}
We employed a suite of baseliens to evaluate on our data, which are: \\
(1) Text-only SLMs: BERT-base-uncased~\cite{devlin2019bert}, DistilBERT~\cite{victor2019distilbert}, RoBERTa-base~\cite{yinhan2019roberta}, GPT-2 (small)~\cite{radford2019language}, and BART-base~\cite{lewis2020bart}.\\
(2) Decoder-only LLMs (zero-/few-shot and IFT): open-source LLaMA-3.2-Instruct~\cite{meta2024llama}, Mistral-7B-Instruct~\cite{mistral20237b}; closed-source: GPT-4o~\cite{gpt4o}, Gemini~1.5~Pro~\cite{team2023gemini}.\\
(3) VLMs (text+image): Phi-3-Vision-128k-Instruct~\cite{phi3mini2024}, LLaVA-1.6~\cite{li2024llava}, LLaMA-3.2-11B-Vision-Instruct~\cite{meta2024llama}, plus GPT-4o-mini~\cite{gpt4o} and Gemini~1.5~Pro~\cite{team2023gemini}.

For text–image annotation we used LLaVA-NeXT~\cite{liu2024llavanext}, Phi-3-Vision-128k-Instruct~\cite{microsoft2024phi}, and MiniCPM-V-2\_6~\cite{yao2024minicpm}, resolving conflicts via a two-step hierarchical voting scheme (details in Appx.~\ref{app:prompts}).

\textbf{Protocols and metrics.}
We evaluated models under three regimes: zero-shot, five-shot, and full fine-tuning. The dataset was partitioned into training, validation, and test splits with an 80/10/10 ratio, and we additionally employed 5-fold cross-validation to assess robustness. For classification tasks, we report standard metrics including precision, recall, F1 score, and accuracy. To capture performance on open-ended outputs, we further evaluate reasoning quality using LLM-as-judge assessments, focusing on reasoning accuracy and faithfulness. We used GPT4o as the judge model.

We compare multimodal models in two complementary setups. In the closed-ended setting, models generate only a categorical label (biased or not biased), and evaluation relies on accuracy-based metrics against ground-truth annotations. In contrast, the open-ended setting (oVQA) requires models to produce both a label and a concise rationale. In this case, rationales are assessed by an LLM judge, which evaluates their correctness and consistency with the given evidence.

\textbf{Exploratory Analysis
}Figure \ref{fig:eda_minipage} presents an overview of our dataset through exploratory data analysis and more in Appendix Figure \ref{fig:eda-onepage}.

\begin{table}[t]
\centering
\caption{\textbf{Vision--Language (Text+Image) Performance.}
Metrics are reported in percentages: Precision (Prec.), Recall, F1, and Accuracy (Acc.). 
Closed-source models are marked with $^\dagger$. Results are grouped into SLMs and VLMs. 
For large models (Phi, LLaVA, LLaMA), we report \textit{Instruct} variants.
Configurations include fine-tuning (FT), instruction fine-tuning (IFT), and few-shot (0-shot, 5-shot).
All models use the same splits and both modalities.}
\footnotesize
\renewcommand{\arraystretch}{1}
\setlength{\tabcolsep}{2pt}
\begin{tabular}{l l *{4}{c}}
\toprule
\textbf{Model} & \textbf{Config} & \textbf{Prec. (\%)} & \textbf{Recall (\%)} & \textbf{F1 Score (\%)} & \textbf{Acc. (\%)} \\
\midrule
\multicolumn{6}{l}{\textit{Small VLMs}} \\
\midrule
\rowcolor{SmallVLMBlue} XLNet + VGG-19 & FT & 72.0 & 68.2 & 70.1 & 77.1 \\
\rowcolor{SmallVLMBlue} BERT + RN-34 & FT & 75.8 & 71.5 & 73.6 & 79.4 \\
\rowcolor{SmallVLMBlue} BERT + CLIP & FT & \textbf{81.3} & \textbf{73.4} & \textbf{77.2} & \textbf{84.2} \\
\rowcolor{SmallVLMBlue} DistilBERT + CLIP & FT & 68.5 & 64.0 & 66.2 & 74.9 \\
\rowcolor{SmallVLMBlue} RoBERTa + CLIP & FT & \textbf{84.5} & \textbf{81.4} & \textbf{82.9} & \textbf{83.6} \\

\midrule
\multicolumn{6}{l}{\textit{Large VLMs}} \\
\midrule
\rowcolor{LargeVLMPeach} Phi-3-Vision & 0-shot & 70.4 & 66.0 & 68.1 & 69.8 \\ 
\rowcolor{LargeVLMPeach}
 & 5-shot & 73.2 & 71.0 & 72.1 & 70.5 \\ 
 \rowcolor{LargeVLMPeach}
 & IFT & 76.8 & 78.1 & 77.4 & 74.0 \\ 
 \rowcolor{LargeVLMPeach}
LLaVA-1.6 & 0-shot & 62.5 & 60.8 & 61.6 & 62.7 \\ \rowcolor{LargeVLMPeach}
 & 5-shot & 68.1 & 67.0 & 67.5 & 65.2 \\ 
 \rowcolor{LargeVLMPeach}
 & IFT & 75.4 & 74.6 & 75.0 & 76.1 \\ 
 \rowcolor{LargeVLMPeach}
LLaMA-3.2-11B & 0-shot & 65.0 & 68.9 & 66.9 & 68.4 \\ \rowcolor{LargeVLMPeach}
 & 5-shot & 73.4 & 74.2 & 73.8 & 72.1 \\
\midrule
\rowcolor{LargeVLMSand} 
GPT-4o mini$^\dagger$ & 0-shot & 71.8 & 74.6 & 73.2 & 72.9 \\ \rowcolor{LargeVLMSand}
 & 5-shot & 77.9 & 79.8 & 78.8 & 77.2 \\ 
 \rowcolor{LargeVLMSand}
Gemini-1.5 Pro$^\dagger$ & 0-shot & 70.9 & 73.1 & 72.0 & 71.5 \\ \rowcolor{LargeVLMSand}
 & 5-shot & 76.8 & 78.5 & 77.6 & 76.9 \\
\bottomrule
\end{tabular}
\label{tab:vlm}
\end{table}

\begin{table}[t]
\centering
\small
\caption{Reasoning accuracy and faithfulness (LLM-judged) for open-ended rationales. Closed-source models are marked with $\dag$.}
\begin{tabular}{l l cc}
\toprule
\textbf{Model} & \textbf{Config} & \textbf{Reasoning Acc. (\%)} & \textbf{Faithfulness (\%)} \\
\midrule
 \rowcolor{LargeVLMPeach}
Phi-3-Vision & 0-shot & 58.2 & 71.0 \\
 \rowcolor{LargeVLMPeach}
             & 5-shot & 62.7 & 74.5 \\
 \rowcolor{LargeVLMPeach}
             & IFT    & 67.9 & 79.8 \\
\rowcolor{LargeVLMPeach}
LLaVA-1.6    & 0-shot & 52.4 & 68.1 \\
\rowcolor{LargeVLMPeach}
             & 5-shot & 55.8 & 70.2 \\
\rowcolor{LargeVLMPeach}
             & IFT    & 68.5 & 80.4 \\
\rowcolor{LargeVLMPeach}
LLaMA-3.2-11B & 0-shot & 59.6 & 72.3 \\
\rowcolor{LargeVLMPeach}
              & 5-shot & 65.4 & 78.0 \\
\midrule
\rowcolor{LargeVLMSand}
GPT-4o mini $\dag$ & 0-shot & 65.1 & 80.2 \\
\rowcolor{LargeVLMSand}
                   & 5-shot & 70.8 & 84.6 \\
\rowcolor{LargeVLMSand}
Gemini-1.5 Pro $\dag$ & 0-shot & 63.4 & 79.0 \\
\rowcolor{LargeVLMSand}
                      & 5-shot & 70.2 & 83.1 \\
\bottomrule
\end{tabular}
\label{tab:reasoning_all}
\end{table}
\subsection{Performance Evaluation}
\label{sec:results}
In this section, we present the performance of multimodals on our dataset and show main results.

\subsubsection{Traditional fine-tuning is still competitive for classification}
We compare different VLMs  (both small vs. large) and across regimes (0-shot, 5-shot, IFT) in a multimodal setting and present results in Table \ref{tab:vlm}.
The results in Table ~\ref{tab:vlm} show clear trends across scales and regimes. Among small VLMs, RoBERTa+CLIP attains the best F1 (82.9) and Acc (83.6), followed by BERT+CLIP (77.2 F1), highlighting gains from strong language encoders with multimodal backbones. Large open-source models improve from 0-/5-shot to IFT (e.g., Phi-3-Vision from 68.1 to 77.4 F1; LLaVA-1.6 from 61.6 to 75.0 F1). Closed-source systems perform strongly in 5-shot (e.g., GPT-4o mini 78.8 F1, Gemini-1.5 Pro 77.6 F1). \\
Overall, our result findings show: (i) CLIP boosts small-model performance, (ii) IFT markedly improves large open-source models, and (iii) closed-source instruct models remain competitive.

\subsubsection{Open-ended reasoning (oVQA):  Strong classifiers also tend to be strong Reasoners }
In the open-ended setting (oVQA), we evaluate autoregresive VLM models output both a label and a short rationale. We then use a judge LLM (GPT4o) to evaluate the rationale, reporting reasoning accuracy (the rate at which the rationale correctly supports the ground-truth decision) and faithfulness (the rate at which the rationale is supported by the given input).  We also measure the correlation between closed-ended accuracy and open-ended reasoning accuracy across all models and regimes.

\begin{figure}[h]
  \centering
  \input{figures/pearson}
  \caption{Closed vs open-ended reasoning accuracy across VLMs. 
  The dashed diagonal indicates perfect alignment ($y=x$). 
  Closed-source models (red triangles) cluster higher than open-source models (blue circles).}
  \label{fig:pearson}
\end{figure}

The results in Table~\ref{tab:reasoning_all} show that reasoning accuracy is consistently lower than closed-ended accuracy, highlighting the difficulty of generating correct and grounded rationales compared to predicting labels. Across open-source models, reasoning accuracy ranges from 52--68\% in the 0- and 5-shot settings, improving to nearly 69\% under instruction tuning. Faithfulness scores follow a similar upward trend, with most models exceeding 75\% once tuned, indicating that rationales are increasingly supported by the input. Closed-source systems (GPT-4o mini, Gemini-1.5 Pro) achieve higher reasoning accuracy and faithfulness overall, with GPT-4o IFT reaching close to 79\% reasoning accuracy and nearly 89\% faithfulness.  

Despite the performance gap as seen in Table~\ref{tab:reasoning_all}, the correlation analysis (Pearson $r = 0.91$) in Figure \ref{fig:pearson} shows a strong linear relationship between closed-ended accuracy and reasoning accuracy. This suggests that models which are strong classifiers also tend to be strong reasoners, although reasoning lags consistently behind classification by 6--12 percentage points. Instruction tuning narrows this gap for both open- and closed-source models, but does not eliminate it entirely.  \\
Overall, these result findings show that that closed-source VLMs are stronger reasoners, yet \textit{all} models face a consistent reasoning gap compared to their closed-ended performance.

\subsubsection{Ablation Studies}
We analyze three factors that affect performance on text-only and multimodal tasks: (i) training data size, (ii) fine-tuning technique, and (iii) multimodal fusion strategy. 

\paragraph{Ablation 1: Training Data Size}
We examined how training data size affects performance across multiple VLMs: Phi-3-Vision, LLaVA-1.6, and LLaMA-3.2-11B-Vision. Each was trained on progressively larger subsets of the dataset. 
\input{figures/ablation}

Figure~\ref{fig:ablation-multi-vlm} compares three representative VLMs under different training data sizes. All models show consistent improvements as data grows, but with diminishing returns after the half-data mark. Phi-3-Vision achieves the highest overall F1 (82.0\% at full scale), followed by LLaMA-3.2-11B-Vision (80.4\%) and LLaVA-1.6 (78.8\%). Interestingly, LLaVA lags at small scales but narrows the gap as more data is used, suggesting that weaker initial alignment can be partially compensated by scale. Across models, the relative gain from 1/16 to full data is 9–11 points, indicating that larger training sets are particularly beneficial for multimodal reasoning despite architectural differences.

\paragraph{Ablation 2: Fine-Tuning Techniques (Parameter-Efficient Methods)}
We compare full fine-tuning against parameter-efficient methods across three representative VLMs: Phi-3-Vision, LLaVA-1.6, and LLaMA-3.2-11B-Vision. These techniques adjust only a small subset of parameters (e.g., low-rank adapters, quantization-aware LoRA \cite{hu_lora_2021}, lightweight adapters, or prompt embeddings) to reduce compute and memory costs while retaining model performance.

\begin{table*}[h]
\centering
\small
\caption{Comparison of parameter-efficient fine-tuning methods across three VLMs. 
Methods include: Full Fine-Tuning (FT), Low-Rank Adaptation (LoRA) \cite{hu_lora_2021}, 
Quantized LoRA (QLoRA) \cite{dettmers2023qlora}, Adapters \cite{houlsby2019parameterefficient}, 
and Prompt-Tuning \cite{lester2021powerscaleparameterefficientprompt}. 
Bold highlights the best score per model. Trainable parameter percentages and GPU memory usage 
are shown for reference }
\setlength{\tabcolsep}{7pt}
\renewcommand{\arraystretch}{1.15}
\begin{tabular}{l c c | cc | cc | cc}
\toprule
\multirow{2}{*}{\textbf{Method}} & \multirow{2}{*}{\textbf{Train.\%}} & \multirow{2}{*}{\textbf{GPU Mem.}} &
\multicolumn{2}{c|}{\cellcolor{LargeVLMPeach} \textbf{Phi-3-Vision}} & \multicolumn{2}{c|}{\cellcolor{LargeVLMPeach}\textbf{LLaVA-1.6}} & \multicolumn{2}{c}{\cellcolor{LargeVLMPeach}\textbf{LLaMA-3.2-11B-Vision}} \\
 & & &\cellcolor{LargeVLMPeach} F1 &\cellcolor{LargeVLMPeach} Reason. Acc. &\cellcolor{LargeVLMPeach} F1 &\cellcolor{LargeVLMPeach} Reason. Acc. &\cellcolor{LargeVLMPeach} F1 &\cellcolor{LargeVLMPeach} Reason. Acc. \\
\midrule
\rowcolor{gray!5} Full FT       & 100  & 24.0 & \textbf{82.0} & \textbf{67.9} & \textbf{78.8} & \textbf{65.5} & \textbf{80.4} & \textbf{66.1} \\
LoRA   \cite{hu_lora_2021}        & 1.2  & 10.1 & 81.2 & 66.8 & 77.9 & 64.7 & 79.7 & 65.3 \\
\rowcolor{gray!5} QLoRA \cite{dettmers2023qlora}         & 1.2  & 7.4  & 80.9 & 66.5 & 77.6 & 64.2 & 79.4 & 65.1 \\
Adapters \cite{houlsby2019parameterefficient}     & 5.0  & 12.8 & 81.4 & 67.1 & 78.1 & 65.0 & 79.9 & 65.7 \\
\rowcolor{gray!5} Prompt-Tuning \cite{lester2021powerscaleparameterefficientprompt}& 0.1  & 6.9  & 79.8 & 64.9 & 76.2 & 62.8 & 77.8 & 63.5 \\
\bottomrule
\end{tabular}
\label{tab:ablation-ft-multi}
\end{table*}

Table~\ref{tab:ablation-ft-multi} shows that parameter-efficient methods achieve consistent trends across Phi-3-Vision, LLaVA-1.6, and LLaMA-3.2-11B-Vision. In all cases, LoRA and Adapters recover over 97--99\% of the full fine-tuning performance while using less than 5\% of trainable parameters. QLoRA offers nearly identical accuracy to LoRA but with lower memory requirements, confirming its utility for scaling. Prompt-tuning is the lightest-weight option but lags behind by 2--3 points in both F1 and reasoning accuracy. These results indicate that the efficiency–performance trade-off is robust across different multimodal backbones, with LoRA and QLoRA striking the most favorable balance.

\paragraph{Ablation 3:  Performance gains when moving from text-only to text+image modality}

\begin{figure}[h]
  \centering
  \includegraphics[width=\linewidth]{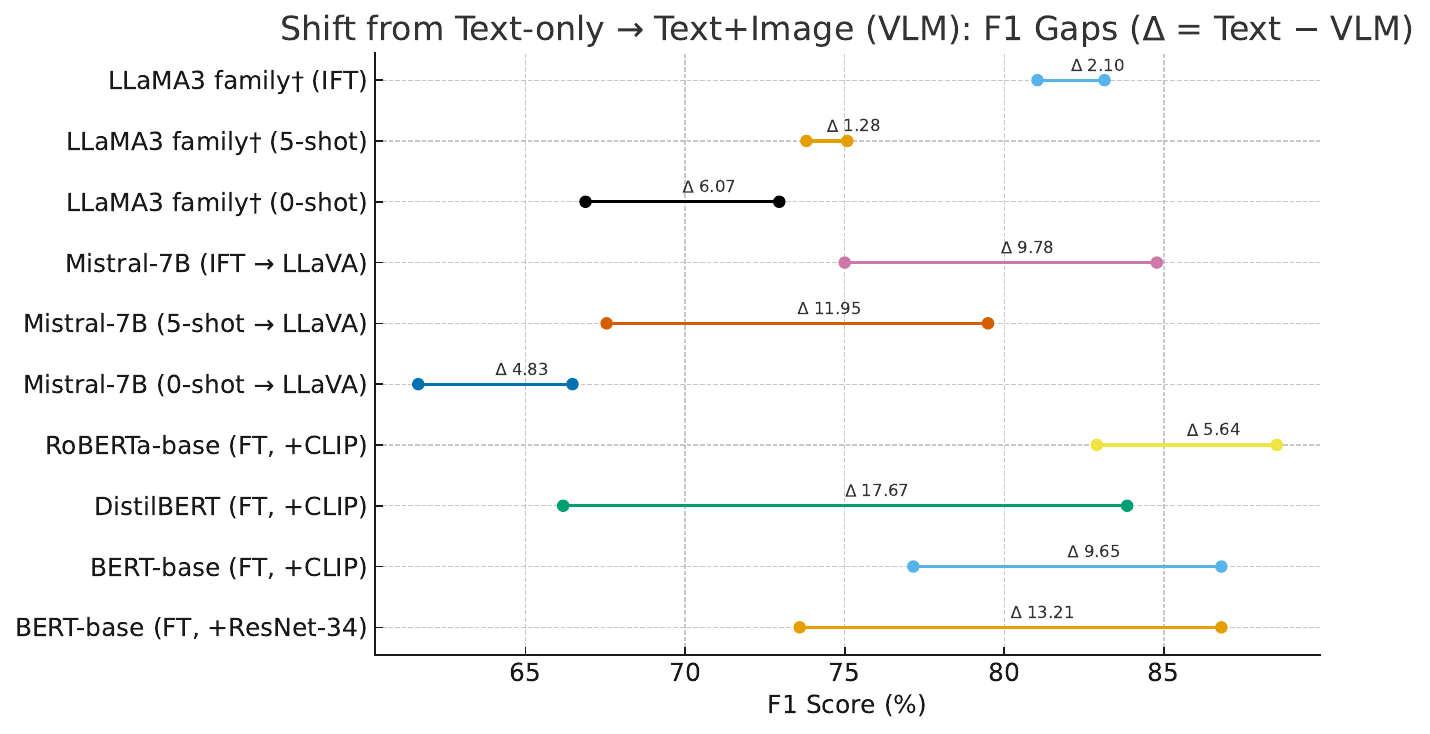}
  \caption{\textbf{Text}\,$\rightarrow$\,\textbf{Text+Image} gap per matched pair.
  Horizontal distance encodes $\Delta{=}\mathrm{F1}_{\text{text}}-\mathrm{F1}_{\text{text+img}}$ (points).
  Gaps decrease with instruction tuning (IFT). $^\dagger$LLaMA3 family aggregates sizes.}
  \label{fig:text2vlm-gap}
\end{figure}

Next, we quantify the effect of adding images by pairing each text-only model/regime with its closest multimodal counterpart and reporting the gap
$\Delta = \mathrm{F1}_{\text{text}} - \mathrm{F1}_{\text{text+img}}$ in Figure \ref{fig:text2vlm-gap}. The results show that across ten matched models pairs, the average gap is \textbf{8.22} points (median \textbf{7.86}; range \textbf{1.28}–\textbf{17.67}) (We do not report proprietary API results in this result to avoid non-reproducible comparisons).
Gaps are largest for smaller backbones and naïve fusion (e.g., DistilBERT+CLIP, $\Delta{=}17.67$) and shrink with instruction tuning (e.g., LLaMA3 family, IFT, $\Delta{=}2.10$). Overall, the results shows that while images can help qualitatively (rationales/grounding), our results show a consistent performance gain when moving from text-only to text+image that narrows as alignment improves.

\paragraph{Qualitative Analysis }
\label{sec:qual_analysis_bias}

\begin{figure}[h]
    \centering
    \includegraphics[width=0.65\linewidth]{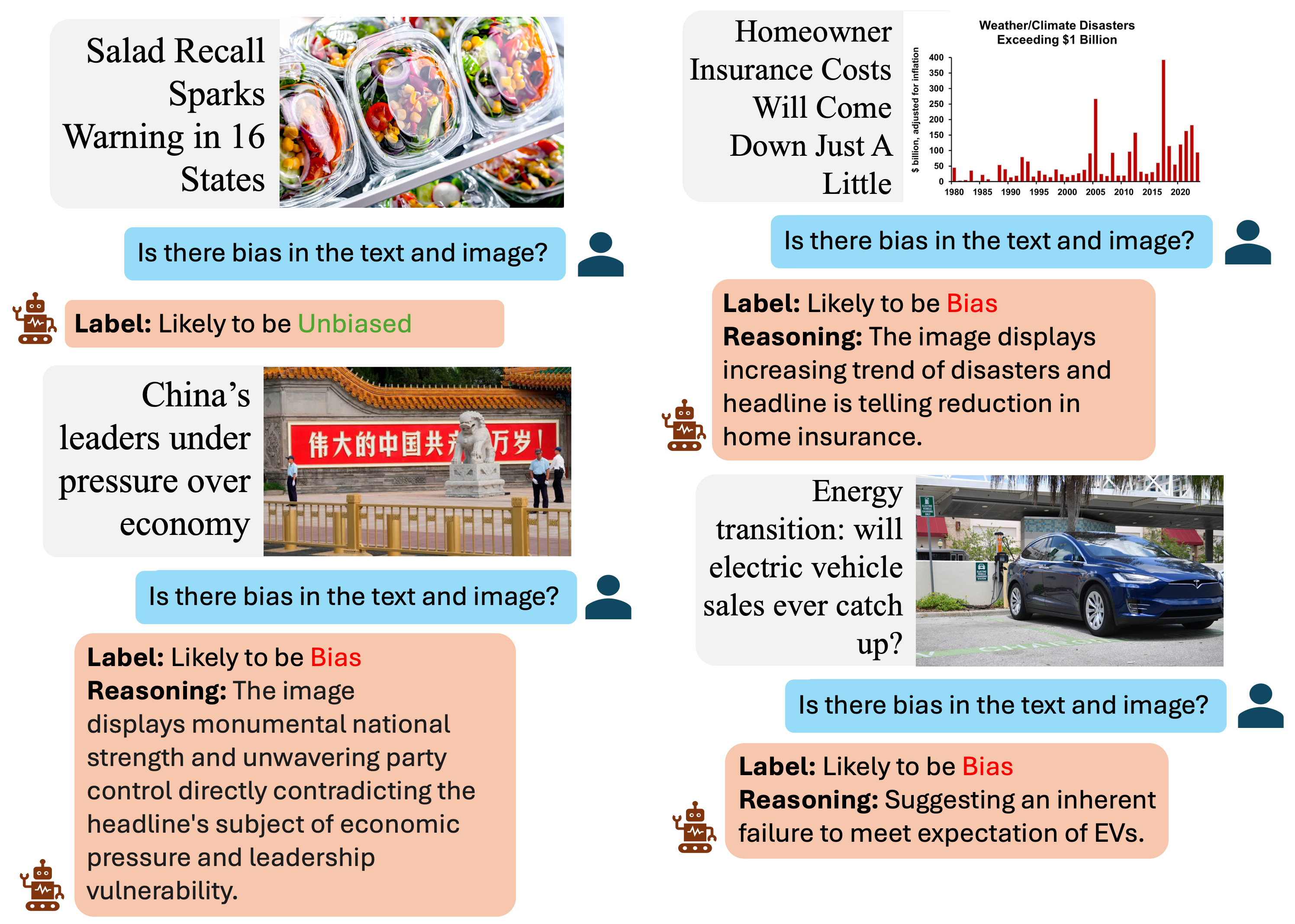}
    \caption{Sample examples from the our dataset showing the reasoning for the biased label.}
    \label{fig:qualitative_results}
\end{figure}

We present qualitative analysis to illustrate how models reason about bias labels. As shown in Figure~\ref{fig:qualitative_results} and Table~\ref{tab:vlm_bias_results_with_images}, models sometimes overgeneralize by flagging bias when captions reference marginalized groups or diversity but the image lacks explicit markers, while others rely heavily on literal textual cues and treat minor mismatches as bias. Although they perform well when text and visuals are clearly aligned, ambiguous or nuanced cases continue to reveal limitations in multimodal reasoning, highlighting the need for richer context modeling in bias detection.

%% file: figures/pearson.tex
\begin{tikzpicture}
  \begin{axis}[
      width=0.7\linewidth,
      height=0.6\linewidth,
      xlabel={Closed Accuracy (\%)},
      ylabel={Reasoning Accuracy (\%)},
      xmin=60, xmax=90,
      ymin=50, ymax=90,
      grid=major,
      legend style={at={(0.5,-0.15)},anchor=north,legend columns=-1},
      legend cell align={left}
    ]
    \addplot[dashed,gray] coordinates {(60,60) (90,90)};

    \addplot+[only marks,blue,mark=o,mark size=2pt] coordinates {
      (69.8,58.2) (70.5,62.7) (74.0,67.9)  
      (62.7,52.4) (65.2,55.8) (76.1,68.5)  
      (68.4,59.6) (72.1,65.4)              
    };
    \addlegendentry{Open-source VLMs}

    \addplot+[only marks,red,mark=triangle*,mark size=2pt] coordinates {
      (72.9,65.1) (77.2,70.8) (86.1,78.9)  
      (71.5,63.4) (76.9,70.2)              
    };
    \addlegendentry{Closed-source VLMs}
  \end{axis}
\end{tikzpicture}

%% file: figures/ablation.tex
\begin{figure}[h]
\centering
\begin{tikzpicture}
\begin{axis}[
  width=0.78\linewidth, height=0.5\linewidth,
  xlabel={\# Training Samples (log scale)}, xmode=log,
  ylabel={F1 Score (\%)}, ymin=50, ymax=85,
  grid=major,
  legend style={at={(0.5,-0.2)}, anchor=north, legend columns=-1},
  legend cell align={left}
]
\addplot+[mark=triangle*,red] coordinates {
  (3100,72.5) (6200,75.9) (12500,78.3) (25000,80.1) (50000,82.0)
};
\addlegendentry{Phi-3-Vision}

\addplot+[mark=o,blue] coordinates {
  (3100,68.2) (6200,71.4) (12500,74.6) (25000,76.5) (50000,78.8)
};
\addlegendentry{LLaVA-1.6}

\addplot+[mark=square*,green] coordinates {
  (3100,69.1) (6200,73.0) (12500,75.8) (25000,78.0) (50000,80.4)
};
\addlegendentry{LLaMA-3.2-11B-Vision}

\end{axis}
\end{tikzpicture}
\caption{Scaling curves of multiple VLMs trained with subsets of increasing size. All models improve with more data, though absolute F1 scores and scaling rates differ.}
\label{fig:ablation-multi-vlm}
\end{figure}

%% file: section/6-Discussion.tex
\section{Discussion}

\subsection{Practical and Social Impact}
Our findings carry both practical and social significance. By applying LLMs and VLMs, we introduce a scalable framework for detecting biases within multimodal media. This is particularly relevant for content moderation, journalism, and public policy, where identifying subtle or underrepresented viewpoints can improve accountability and foster greater transparency.

Socially, our framework empowers more equitable media practices by revealing hidden biases that shape public opinion and perpetuate stereotypes. In practical terms, this means stakeholders can systematically uncover and address harmful or skewed narratives. Integrating human–machine collaboration also allows for efficient annotations without sacrificing critical context or cultural sensitivity, lowering the chances of purely algorithmic errors.

Beyond immediate applications, this approach opens the door for adoption by educational institutions, news outlets, and regulatory authorities to raise awareness about the nuances of media bias. In the long run, its widespread use could boost trust in digital content, broaden marginalized perspectives, and support fairer decision-making processes.

\subsection{Limitations}
Despite promising outcomes, our work comes with certain limitations. First, although our dataset spans a diverse range of media sources, it may not fully capture global cultural or linguistic nuances, which could affect the generalizability of our results. Second, our framework prioritizes biases that are more readily detected via textual or visual cues, potentially missing more pervasive systemic or cultural biases requiring deeper contextual insight.

Additionally, dependence on SoTA language and vision-language models brings its own challenges, mainly significant computational overhead and resource requirements, which could be prohibitive for smaller organizations or those operating in lower-resource settings. While the hybrid human–machine annotation approach improves overall reliability, it is not entirely free from subjective disagreements, especially for ambiguous content  \cite{raza2023entity}.

Media bias is also dynamic, shifting over time. Our dataset covers the period from May 2023 to September 2024, meaning newly emerging trends or evolving patterns may fall outside its scope. Future research should aim to include a broader range of cultures, languages, and emerging media practices, as well as more granular bias categories, while exploring real-time adaptability to keep pace with rapid media developments.

%% file: section/7-Conclusion.tex
\section{Conclusion}
In this study, we introduced \textbf{ViLBias}, a comprehensive framework designed to detect bias in multimodal contexts by capturing both linguistic and visual cues. Our results demonstrate that LLMs and VLMs can effectively identify subtle biases in text and images when supported by an intelligently curated dataset and a robust human–machine hybrid annotation strategy. Together, these efforts advance our goal of improving transparency in media and increasing trust in digital content.
Performance metrics such as F1 score and accuracy confirm the reliability of LLM-based annotations. Furthermore, our majority-voting mechanism mitigates stochastic variations, enhancing the stability of the classification process. We also highlight the benefits of instruction-based fine-tuning and lightweight techniques like LoRA, enabling strong results even in resource-limited environments. Taken together, these contributions show the potential of \textbf{ViLBias}in bias detection in multimodal media scenarios.

%% file: section/8-Appendix.tex
\renewcommand{\thefigure}{A.\arabic{figure}}
\renewcommand{\thetable}{A.\arabic{table}}
\setcounter{figure}{0}
\setcounter{table}{0}
\appendix
\section{Data Collection}
The dataset includes a diverse range of news sources such as The Guardian, CBS News, ABC News, Bloomberg, The Atlantic, and Reuters. Additionally, sources like Fox News, Forbes, CNN, CNBC, and The Hill contribute to a broad spectrum of coverage. Prominent outlets such as The New York Times, Washington Post, Al Jazeera, Time, and The Daily Beast are also included. The dataset incorporates international and regional perspectives from BBC, CBC, Global News, AP News, and The Globe and Mail. Other sources range from Politico, The Economist, The Federalist, and ProPublica to Axios, Newsweek, Daily Kos, and National Review. Conservative-leaning sources such as Newsmax, OANN, Breitbart, Daily Caller, and The Washington Examiner are present, along with more liberal-leaning platforms like HuffPost, New Yorker, and NPR/PBS NewsHour. Financial and economic perspectives are represented through WSJ, FT, and CNBC, while tabloid-style and opinion-driven outlets like New York Post, Toronto Sun, and National Post provide additional viewpoints. This collection ensures a balanced mix of mainstream, independent, and alternative news coverage. \\


\section{Annotation Guidelines}
\label{app:ann}
Annotators were asked to follow these specific guidelines:
\begin{enumerate}
    \item \textbf{Accuracy}: Verify whether the annotation aligns correctly with the provided input (e.g., text or text-vision pair). For text-based tasks, the annotation must correctly reflect the content of the text. For text-vision pairs, the annotation must accurately describe the relationship between the textual and visual content.
    \item \textbf{Completeness}: Ensure that the annotation captures all relevant details without omitting critical information. For instance, descriptions should be neither overly brief nor excessively verbose.
    \item \textbf{Relevance}: Assess whether the annotation remains within the scope of the prompt and does not include extraneous or speculative information.
    \item \textbf{Consistency}: Compare annotations across similar inputs within the sample to ensure uniformity in style and language. Discrepancies in tone, terminology, or format should be flagged.
    \item \textbf{Bias Detection}: Identify any systematic biases, such as stereotypes or skewed representations in text or vision-text pairs, that might emerge from the model's outputs. Annotators were specifically instructed to consider the following bias categories, such as linguistic, visual, gender, racial, cutlural , disbility and age related biases.
    \item \textbf{Tie-Breaking Validation}: For cases where model annotations indicate a tie in voting, verify if the chosen annotation aligns with the overall context and the majority pattern in other cases.
\end{enumerate}
Annotators were instructed to record their evaluations using a structured review form, categorizing annotations as \textit{Acceptable}, \textit{Needs Improvement}, or \textit{Incorrect}, with detailed comments explaining their reasoning. The review also included a quantitative agreement score between human judgments and model-generated annotations to measure annotation quality. The feedback collected was used to refine model prompts and guide further improvements in the annotation pipeline.

\begin{table}[h]
\footnotesize
\centering
\caption{Key settings for training, generation, and evaluation.}
\begin{tabular}{R{0.22\linewidth}L{0.66\linewidth}}
\toprule
\textbf{Generation} & Temperature 0.2–0.4; max tokens 1024 \\ \midrule
\textbf{Batch / LR} & BERT: 32, 15 epochs, $2\!\times\!10^{-5}$, wd 0.01, ES patience 5 \\
                    & FT (VLM/LLM): batch 1, 3 epochs, $2\!\times\!10^{-4}$, grad acc 1, max grad norm 0.3 \\ \midrule
\textbf{Scheduler}  & Constant LR with 0.03 warmup ratio \\ \midrule
\textbf{Precision}  & Mixed precision (\texttt{bf16=True}) \\ \midrule
\textbf{Loss}       & CrossEntropy; Focal (\(\alpha{=}1,\ \gamma{=}2\)) where noted \\ \midrule
\textbf{Eval}       & Accuracy, Precision, Recall, F1; confusion matrices (batch 32) \\ \midrule
\textbf{Quant.}     & 4-bit (\texttt{nf4}) with double quantization \\ \midrule
\textbf{Reproduc.}  & CUDA devices (CPU fallback), seed 0; \texttt{TOKENIZERS\_PARALLELISM=false} \\ 
\bottomrule
\end{tabular}
\label{tab:training_params}
\end{table}

\section{Prompts}
\label{app:prompts}
\begin{tcolorbox}[breakable, width=1.0\linewidth,
                  title=Prompt for Text-based Annotation,
                  colframe=black!70, colback=white, fonttitle=\bfseries]
\scriptsize
\begin{lstlisting}[language=Matlab, breaklines=true]
prompt = f"""
    Analyze the provided text for evidence of bias. Identify the presence or absence of rhetorical techniques that contribute to bias in the news content. Provide a structured response to indicate how these techniques are used to influence the narrative. 
    Text: {text}

    Rhetorical Techniques to Identify:
    1. Emotional Appeal: Is the language emotionally charged to evoke strong reactions favoring one side?
    2. Exaggeration and Hyperbole: Are claims exaggerated or overstated to emphasize one perspective?
    3. Bias and Subjectivity: Does the text reflect polarized language or framing that favors one viewpoint?
    4. Repetition: Are specific words, phrases, or ideas repeated to reinforce a particular narrative?
    5. Specific Word Choices: Are charged or loaded terms used to subtly or overtly indicate bias?
    6. Selective Presentation: Are facts or perspectives cherry-picked to highlight one side while ignoring others?
    7. Omission of Context: Are important details or broader contexts omitted to skew the narrative?
    8. Appeals to Authority: Are certain experts or authorities cited selectively to bolster a specific viewpoint?
    9. Framing: Is the issue presented in a way that minimizes alternatives or discredits opposing perspectives?
    10. Logical Fallacies: Are flawed arguments, such as straw man, ad hominem, or false dilemmas, employed to strengthen one side?
    11. Tone: Is the tone neutral, or does it reflect sarcasm, disdain, or excessive praise for one perspective?
    12. Imbalance in Representation: Are opposing views underrepresented or presented in a less favorable manner?
    13. Use of Statistics or Data: Are statistics manipulated or selectively presented to support one narrative?

    Provide your analysis in the following format:
    - Emotional Appeal: [Present/Absent]
    - Exaggeration and Hyperbole: [Present/Absent]
    - Bias and Subjectivity: [Present/Absent]
    - Repetition: [Present/Absent]
    - Specific Word Choices: [Present/Absent]
    - Selective Presentation: [Present/Absent]
    - Omission of Context: [Present/Absent]
    - Appeals to Authority: [Present/Absent]
    - Framing: [Present/Absent]
    - Logical Fallacies: [Present/Absent]
    - Tone: [Neutral/Biased]
    - Imbalance in Representation: [Present/Absent]
    - Use of Statistics or Data: [Present/Absent]

    Keep your evaluation concise and structured, without additional explanations.
"""
\end{lstlisting}
    \label{fig:text-prompt}
    \end{tcolorbox}
    \hfill
    
\begin{tcolorbox}[breakable, width=1.0\linewidth,
                      title=Prompt for Text and Image-based Annotation,
                      colframe=black!70, colback=white, fonttitle=\bfseries]
    \scriptsize
\begin{lstlisting}[language=Matlab, breaklines=true]
prompt = f"""
    Analyze the provided text and associated image for bias. Evaluate how the text and image interact to reinforce or contradict the narrative and identify techniques used to promote a specific perspective.

    Text: {text}
    Image Description: {image_description}

    Evaluation Criteria:
    1. Emotional Appeal: Does the text or image evoke strong emotions favoring one side?
    2. Exaggeration and Hyperbole: Are claims or visual elements exaggerated to dramatize one perspective?
    3. Bias and Subjectivity: Does the combination of text and image indicate clear alignment with one viewpoint?
    4. Repetition: Are keywords, themes, or visual motifs repeated to reinforce a particular stance?
    5. Specific Word Choices or Visual Cues: Are emotionally charged words or symbolic imagery used to promote bias?
    6. Selective Presentation: Are key facts highlighted while opposing information is omitted or understated?
    7. Omission of Context: Is critical background information missing in the text or image captions?
    8. Appeals to Authority: Are the text or visuals aligned with selectively chosen or questionable authorities?
    9. Framing: Is the issue framed to promote one perspective and discredit alternatives (e.g., through cropping, captions)?
    10. Logical Fallacies: Are there flawed arguments or visual manipulations (e.g., misleading captions, doctored images)?
    11. Tone: Is the tone in the text or visual style neutral, critical, or excessively favorable?
    12. Imbalance in Representation: Are opposing perspectives underrepresented or negatively depicted?
    13. Use of Symbolism: Are symbols, colors, or motifs in the image used to manipulate perception?
    14. Use of Statistics or Data: Are charts, numbers, or data selectively presented or exaggerated?

    Provide your analysis in the following format:
    - Emotional Appeal: [Present/Absent]
    - Exaggeration and Hyperbole: [Present/Absent]
    - Bias and Subjectivity: [Present/Absent]
    - Repetition: [Present/Absent]
    - Specific Word Choices or Visual Cues: [Present/Absent]
    - Selective Presentation: [Present/Absent]
    - Omission of Context: [Present/Absent]
    - Appeals to Authority: [Present/Absent]
    - Framing: [Present/Absent]
    - Logical Fallacies: [Present/Absent]
    - Tone: [Neutral/Biased]
    - Imbalance in Representation: [Present/Absent]
    - Use of Symbolism: [Present/Absent]
    - Use of Statistics or Data: [Present/Absent]

    Keep your evaluation precise and avoid unnecessary explanations.
"""
\end{lstlisting}
    \label{fig:text-image-prompt}
    \end{tcolorbox}

{\scriptsize
\begin{tcolorbox}[breakable,
    title={LLM-as-Judge Prompt (ViLBias metrics: Reasoning Accuracy \& Faithfulness)},
    colback=white,
    colframe=black!35,
    arc=2mm,
    left=1mm,
    right=1mm,
    top=1mm,
    bottom=1mm]

You are an impartial judge. Score a model’s open-ended rationale for a news \textbf{text+image} item on two binary metrics:
\textbf{(A) Reasoning Accuracy} — does the rationale, if taken as true, \emph{correctly support the ground-truth label}? 
\textbf{(B) Faithfulness} — is the rationale \emph{grounded in the provided input} (text and/or image) with no hallucinated facts.

\textbf{Inputs}
- \texttt{QUESTION}: “Does this content exhibit bias? Please explain your reasoning.”
- \texttt{TEXT}: <<<\{TEXT SNIPPET\}>>>
- \texttt{IMAGE}: (rendered alongside; if unavailable, write “none”)
- \texttt{GROUND\_TRUTH\_LABEL}: \{\texttt{biased} | \texttt{not\_biased}\}
- \texttt{MODEL\_PRED\_LABEL}: \{\texttt{biased} | \texttt{not\_biased}\}
- \texttt{MODEL\_RATIONALE}: <<<\{FREE-TEXT RATIONALE\}>>>

\textbf{Judging Rules}
1) \textbf{Reasoning Accuracy (0/1)}: 
   Assign 1 if the rationale’s claims, taken at face value, \emph{logically entail or strongly support} the \texttt{GROUND\_TRUTH\_LABEL}. 
   Penalize if it argues for the opposite label, is purely label restatement with no reason, or uses irrelevant logic.

2) \textbf{Faithfulness (0/1)}: 
   Assign 1 only if the rationale’s cited cues are \emph{verifiably present} in \texttt{TEXT} and/or \texttt{IMAGE}. 
   Disallow unstated assumptions, world knowledge, or fabricated details. 
   When possible, quote minimal \textbf{text spans} and/or describe \textbf{visual evidence} (objects/attributes/relations); if no image, judge solely on text.

3) \textbf{Edge Cases}: 
   - Empty, generic, or circular rationales (“it is biased because it is biased”) $\rightarrow$ \texttt{ReasoningAccuracy}=0, \texttt{Faithfulness}=0.
   - Correct label but incorrect/unsupported reasons $\rightarrow$ \texttt{ReasoningAccuracy}=0 or \texttt{Faithfulness}=0 as appropriate.
   - If \texttt{MODEL\_PRED\_LABEL} differs from \texttt{GROUND\_TRUTH\_LABEL}, you still judge the \emph{rationale} against the \emph{ground truth}.

\textbf{Output (strict JSON)}
\begin{verbatim}
{
  "reasoning_accuracy": 0 or 1,
  "faithfulness": 0 or 1,
  "justification": "2-3 sentences: why each score was assigned"
}
\end{verbatim}

\textbf{Grading Hints}
- Typical textual bias cues: loaded wording, one-sided framing, selective emphasis/omission. 
- Typical visual bias cues: image choice, cropping/staging, emotive imagery, text–image mismatch. 
- Keep judgments short, precise, and grounded only in the provided inputs.

\end{tcolorbox}

\begin{center}
\captionof{figure}{LLM-as-Judge prompt for evaluation}
\label{fig:judge-prompt}
\end{center}
}


\begin{table}
\centering
\scriptsize
\renewcommand{\arraystretch}{0.8} 
\setlength{\tabcolsep}{2pt} 
\caption{Binary classification sample results for various VLMs.}
\begin{tabular}{@{}p{3.5cm}p{4cm}@{}}
\toprule
\textbf{Model} & \textbf{Input and Analysis} \\
\midrule
\textbf{Llama3.2-vision-11B} \newline
\includegraphics[width=0.18\textwidth]{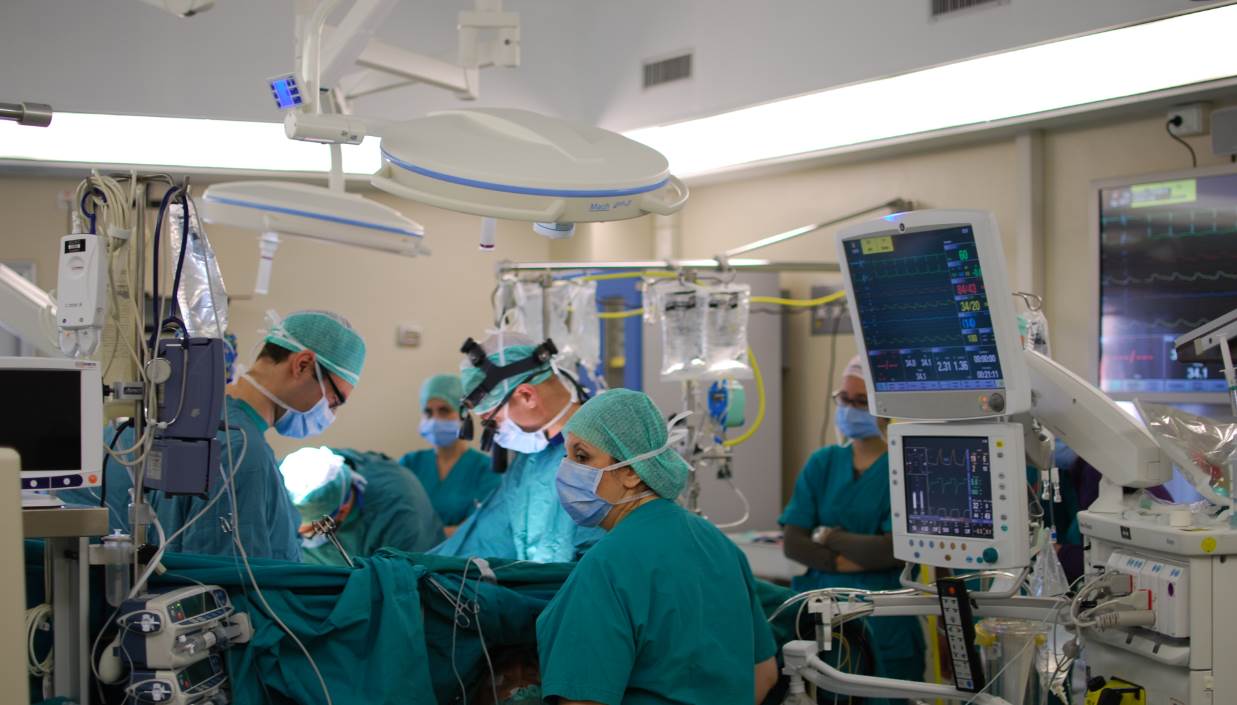} &
\textbf{Caption:} ``A program empowering indigenous women to advocate for their rights.'' \newline
\textbf{Reasoning:} The text highlights empowerment, but the image lacks visible advocacy efforts, suggesting underrepresentation. \newline
\textbf{Bias Observed:} The model assumes that the absence of explicit visual markers is equal to bias, which may be an overgeneralization. \\
\midrule

\textbf{Llama3.2-vision-11B} \newline
\includegraphics[width=0.18\textwidth]{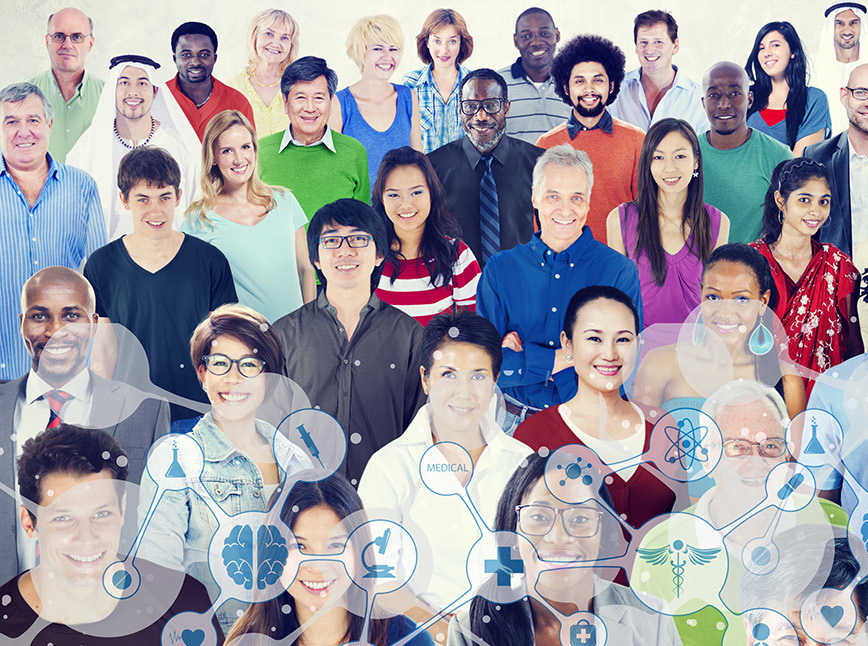} &
\textbf{Caption:} ``A collaborative community event highlighting diverse voices in science.'' \newline
\textbf{Reasoning:} The text emphasizes diversity, but the image lacks representation of marginalized groups, indicating bias. \newline
\textbf{Bias Observed:} The model overrelies on textual implications of diversity and fails to interpret the broader visual context. \\
\midrule

\textbf{Phi-3 Vision} \newline
\includegraphics[width=0.18\textwidth]{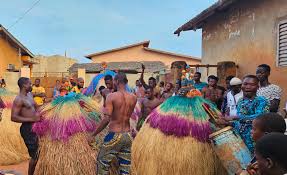} &
\textbf{Caption:} ``A vibrant cultural festival in Africa.'' \newline
\textbf{Reasoning:} The image does not convey the vibrancy described in the text, suggesting bias. \newline
\textbf{Bias Observed:} The model overemphasizes textual cues without fully considering the broader cultural context depicted in the image. \\
\midrule

\textbf{Phi-3 Vision} \newline
\includegraphics[width=0.18\textwidth]{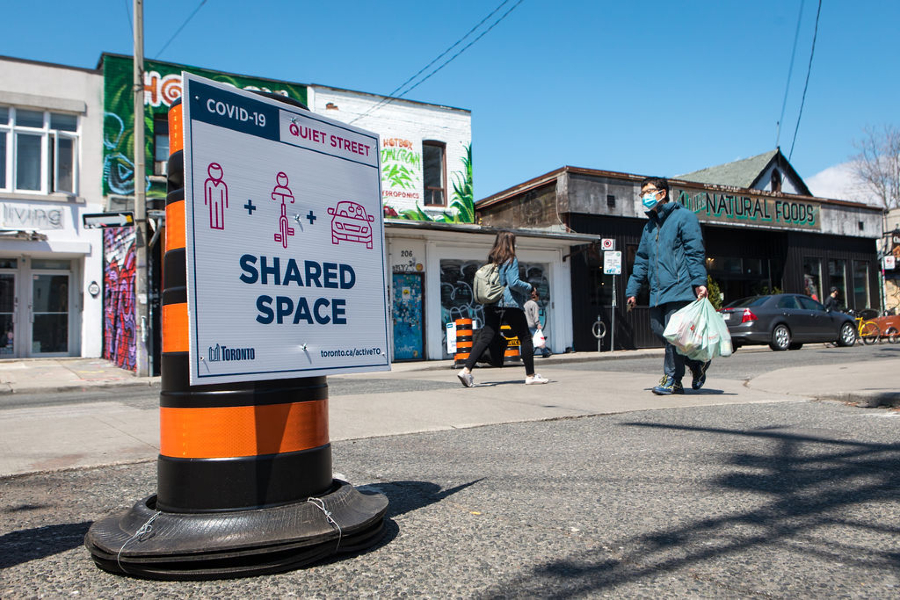} &
\textbf{Caption:} ``A bustling shared street in Toronto during COVID.'' \newline
\textbf{Reasoning:} While the image aligns with the description, certain social cues (e.g., accessibility) are missing, implying bias. \newline
\textbf{Bias Observed:} The model fails to account for nuanced social elements, such as inclusivity markers and diverse pedestrian usage. \\
\midrule

\textbf{Llama3.2-vision-11B} \newline
\includegraphics[width=0.18\textwidth]{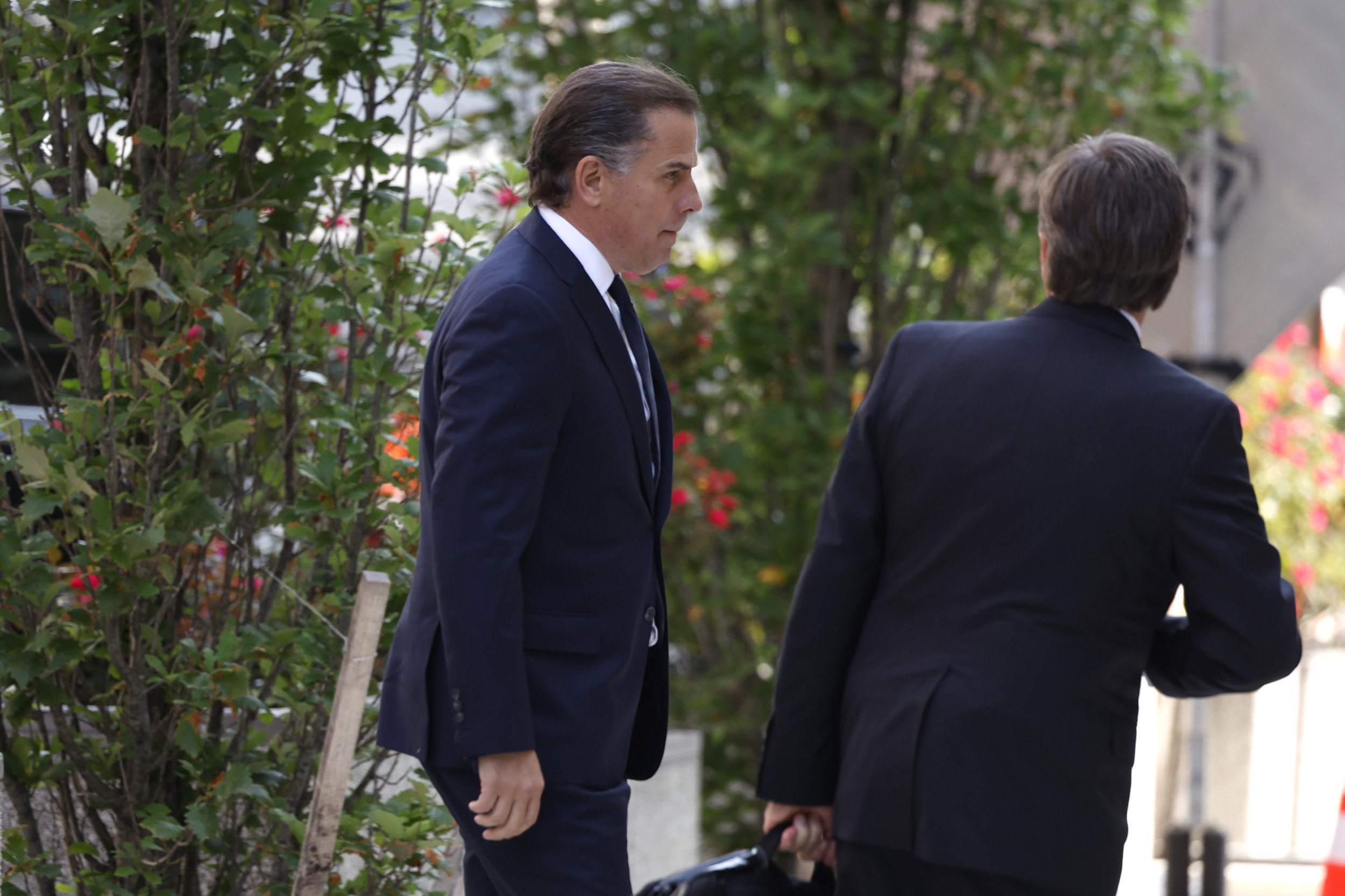} &
\textbf{Caption:} ``Two individuals arrive at a runway for a formal event.'' \newline
\textbf{Reasoning:} The individuals are dressed formally and wearing sunglasses at an airport. Nothing implies a stereotype or underrepresentation. \newline
\textbf{Bias Observed:} \textbf{No bias observed.} The model’s reasoning is correct and neutral—there is no mismatch between the caption and the image. \\
\midrule

\textbf{Phi-3 Vision} \newline
\includegraphics[width=0.18\textwidth]{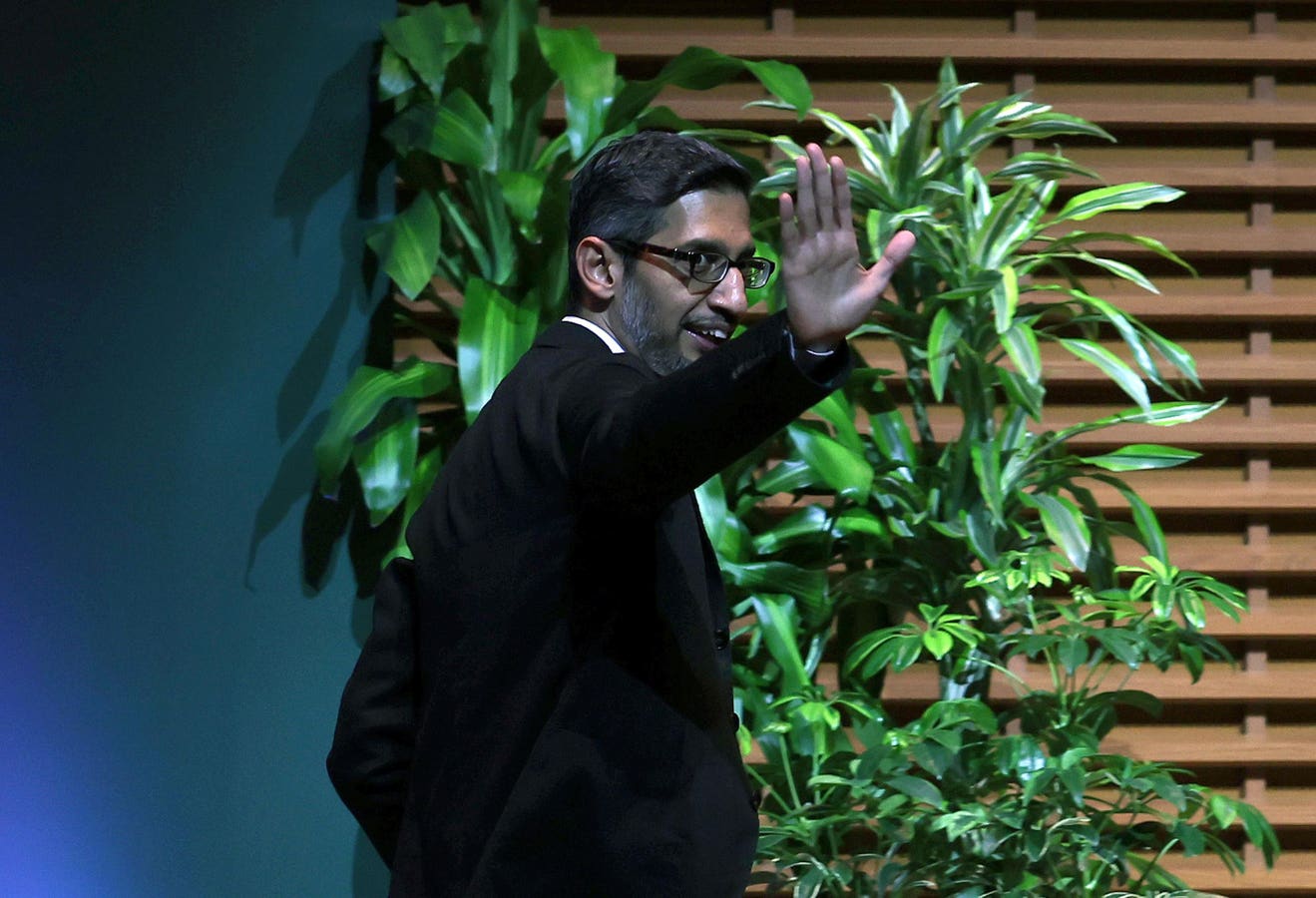} &
\textbf{Caption:} ``A speaker waving to an audience at a tech conference.'' \newline
\textbf{Reasoning:} The person is wearing business attire and appears to be greeting viewers. The caption is consistent with the visual context. \newline
\textbf{Bias Observed:} \textbf{No bias observed.} The model recognizes a straightforward scenario—no overgeneralization regarding the speaker’s background or role. \\

\bottomrule
\end{tabular}
\label{tab:vlm_bias_results_with_images}
\end{table}

\clearpage

%% file: section/haad_eda.tex
\begin{figure*}[h]
    \centering
    \begin{subfigure}{0.42\textwidth}
        \centering
        \includegraphics[width=\linewidth]{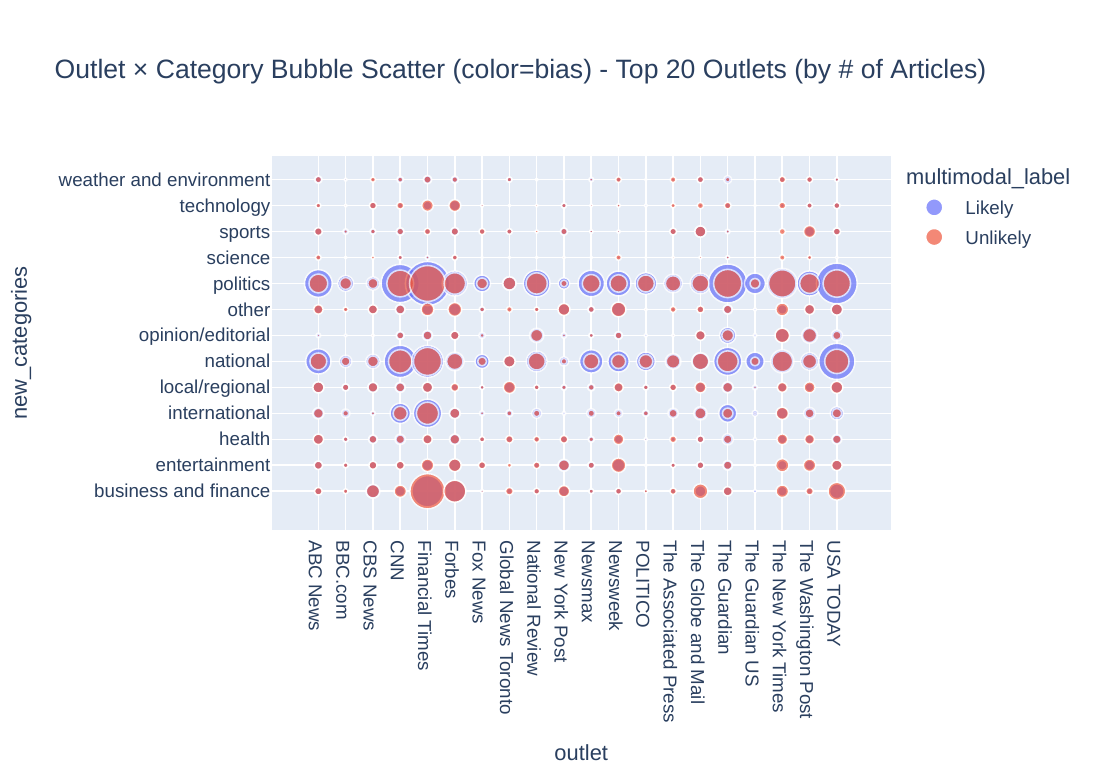}
        \caption{Categories covered by some news outlets. Color represents bias. Size of marker represents the number of articles which have the same newspaper, category and bias.}
        \label{fig:placeholder5}
    \end{subfigure}
    \hfill
    \begin{subfigure}{0.35\textwidth}
        \centering
        \includegraphics[width=\linewidth]{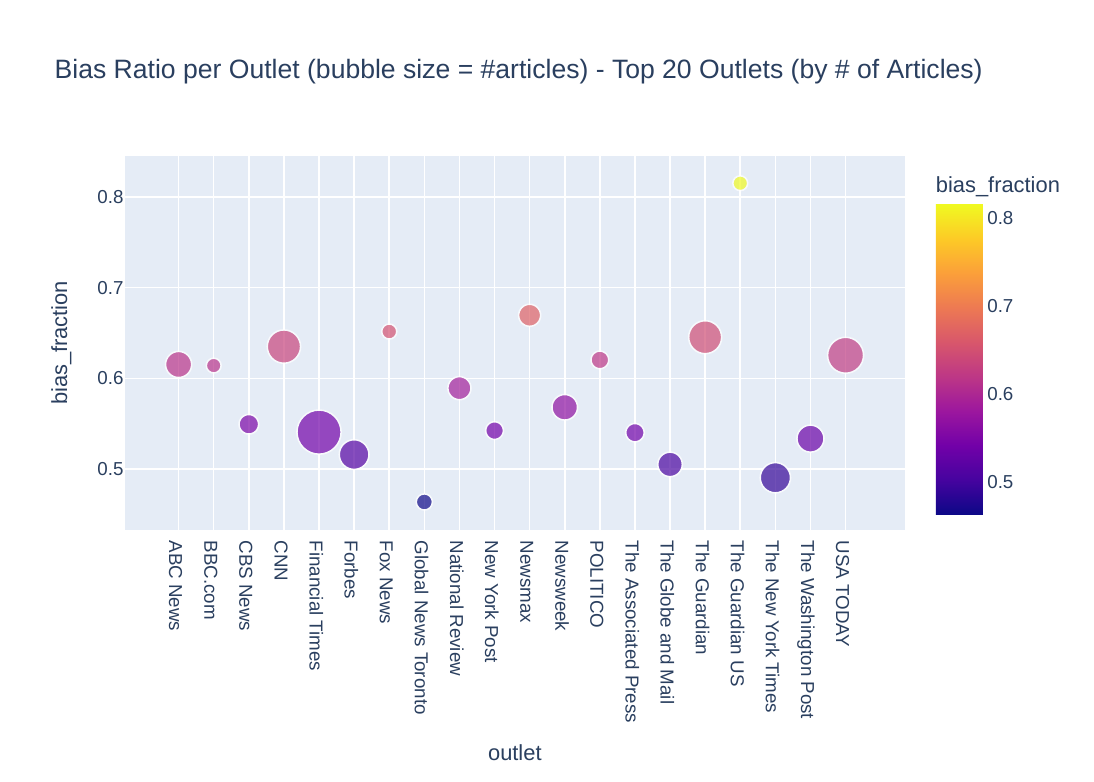}
        \caption{News outlet and the ratio of biased articles that are released. Size of marker represents sample size of articles for that news outlet.}
        \label{fig:placeholder6}
    \end{subfigure}

\vspace{-0.5em}

    \begin{subfigure}{0.35\textwidth}
        \centering
        \includegraphics[width=\linewidth]{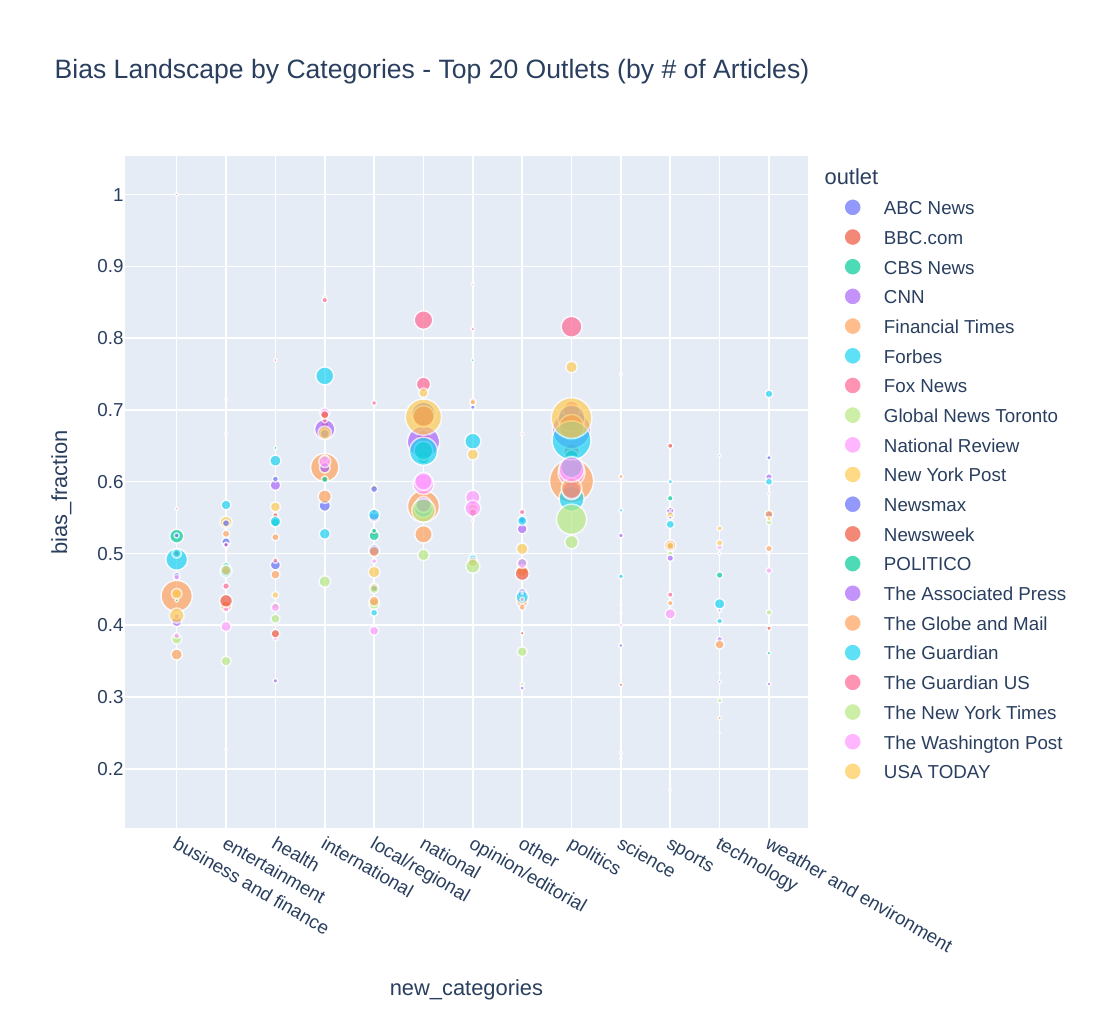}
        \caption{Outlet-wise distribution of bias for each category. Size of marker represents sample size. Vertical position represents the amount of bias of a newspaper in that category.}
        \label{fig:placeholder8}
    \end{subfigure}

    \caption{Distribution charts of bias counts and ratios by outlets and categories. }
    \label{fig:all_figures}
\end{figure*}

\begin{figure*}[h]
    \centering
    \begin{subfigure}{0.42\textwidth}
        \centering
        \includegraphics[width=\linewidth]{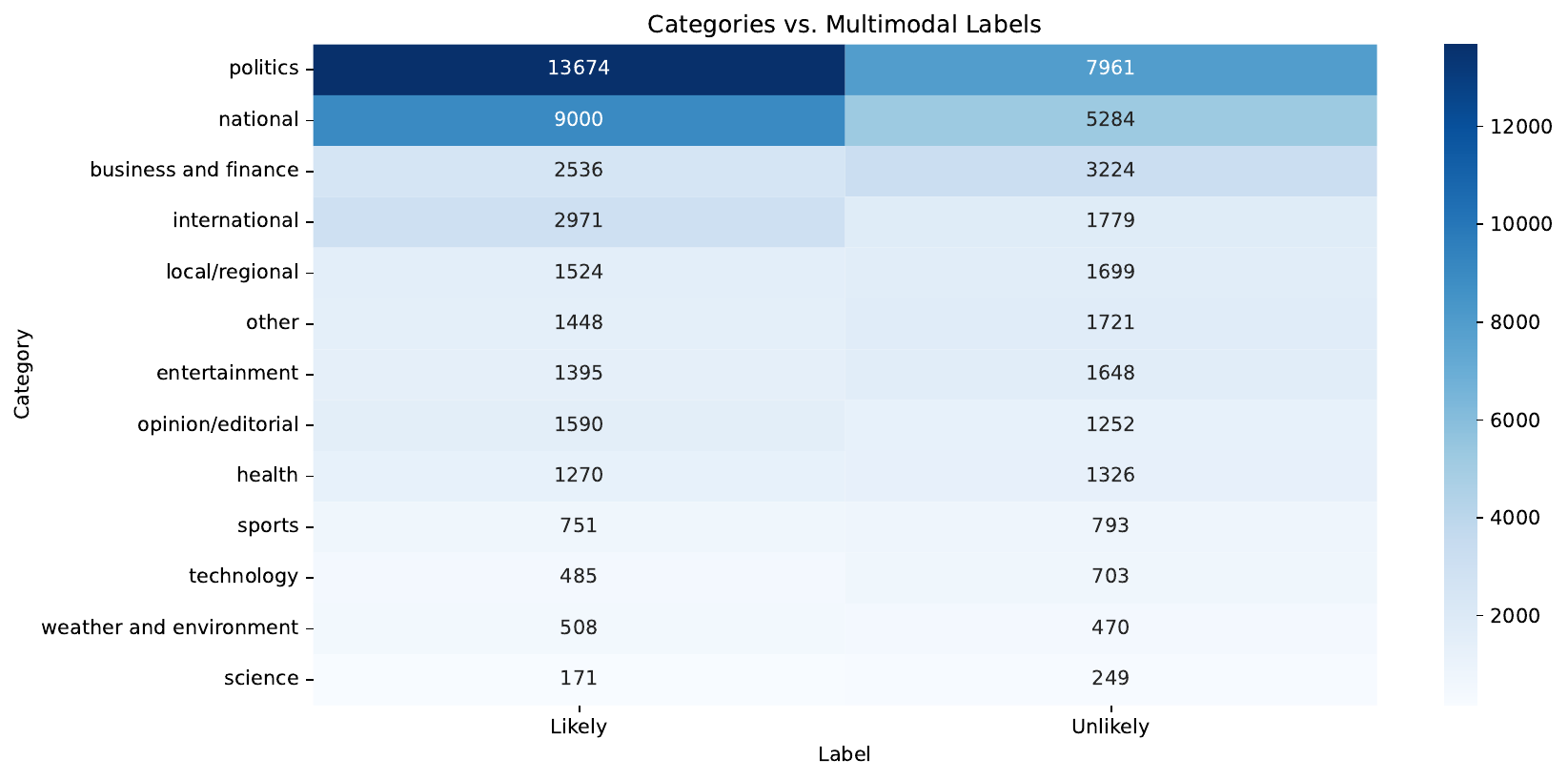}
        \caption{Count of biased articles in every category. Politics is the most affected.}
        \label{fig:placeholder1}
    \end{subfigure}
    \hfill
    \begin{subfigure}{0.42\textwidth}
        \centering
        \includegraphics[width=\linewidth]{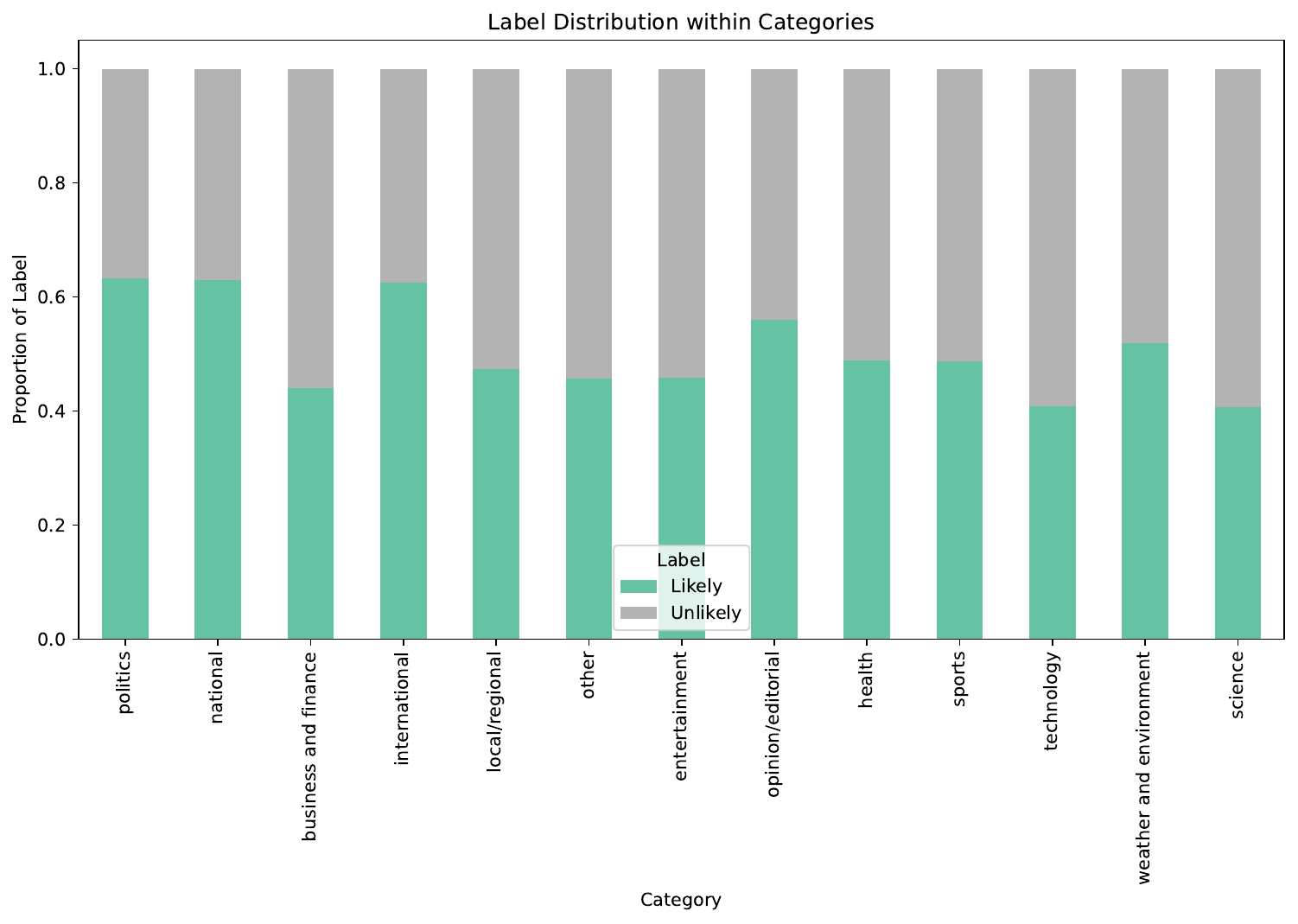}
        \caption{Percentage of biased articles Per category.}
        \label{fig:placeholder2}
    \end{subfigure}
    \caption{Count and percentage-wise distribution of bias per category.}
    \label{fig:all_figures}
\end{figure*}